\pdfoutput=1
\documentclass[11pt,a4paper]{article}
\usepackage[]{acl}

\usepackage{newtxtext}
\usepackage{newtxmath}
\usepackage{latexsym}
\usepackage{anyfontsize}

\usepackage[T1]{fontenc}
\usepackage[utf8]{inputenc}
\usepackage[nopatch=footnote]{microtype}
\usepackage{inconsolata}

\usepackage{graphicx}
\usepackage{enumitem}
\usepackage{float}
\usepackage{subcaption}
\usepackage{multirow}
\usepackage{colortbl}
\usepackage{capt-of}
\usepackage{booktabs}
\usepackage{tabularx}
\usepackage{floatpag}
\usepackage{linguex}
\usepackage{bbding}
\usepackage{cleveref}
\usepackage{siunitx}
\usepackage[section]{placeins}
\usepackage{enumitem}
\usepackage{titlefoot}

\newcolumntype{Y}{>{\centering\arraybackslash}X}

\usepackage[absolute]{textpos}
\usepackage{color}
\definecolor{gray}{gray}{0.5}
\usepackage{helvet}

\title{The Illusion of Competence: Evaluating the Effect of Explanations on Users' Mental Models of Visual Question Answering Systems}

\author{
    \textbf{Judith Sieker$^\dagger$,} 
    \textbf{Simeon Junker$^\dagger$,}
    \textbf{Ronja Utescher$^\dagger$,}
    \textbf{Nazia Attari$^\dagger$,}\\
    \textbf{Heiko Wersing$^\ddagger$,}
    \textbf{Hendrik Buschmeier$^\parallel$,}
    \textbf{Sina Zarrieß$^\dagger$}
\\
    $^\dagger$Computational Linguistics, Department of Linguistics, Bielefeld University\\
    $^\ddagger$Honda Research Institute Europe\\
    $^\parallel$Digital Linguistics Lab, Department of Linguistics, Bielefeld University
}

\begin{document}
\maketitle
\unmarkedfntext{%
    Code and data of study are available at \href{https://doi.org/10.17605/OSF.IO/4KDB5}{https://doi.org/ 10.17605/OSF.IO/4KDB5} or \href{https://github.com/clause-bielefeld/IllusionOfCompetence-VQA-Explanations}{https://github.com/clause-bielefeld/IllusionOfCompetence-VQA-Explanations}.
}

\begin{abstract}
    We examine how users perceive the limitations of an AI system when it encounters a task that it cannot perform perfectly and whether providing explanations alongside its answers aids users in constructing an appropriate mental model of the system's capabilities and limitations. We employ a visual question answer and explanation task
    where we control the AI system's limitations by manipulating the visual inputs: during inference, the system either processes full-color or grayscale images. Our goal is to determine whether participants can perceive the limitations of the system. We hypothesize that explanations will make limited AI capabilities more transparent to users. However, our results show that explanations do not have this effect. Instead of allowing users to more accurately assess the limitations of the AI system, explanations generally increase users' perceptions of the system's competence -- regardless of its actual performance.
\end{abstract}


\section{Introduction}
\label{sec:introduction}

Machine learning-based technologies (often called ‘artificial intelligence’, AI) are now commonly being deployed and used in real-world applications, influencing human decision-making (or automating decision-making altogether) with implications for societies, organizations, and individuals. Despite continuous advances and impressive performance on many tasks, these technologies are not always accurate and will likely never be. Machine learning models depend on curation of the data they are trained on, they are optimized according to criteria that may not do justice to the complexity of reality, and the context in which they are used cannot be fully modeled, to name a few reasons for their limitations. In addition, the underlying algorithms themselves have inherent weaknesses. Large language models (LLMs), e.g., are well known to hallucinate, i.e., to make predictions that are inconsistent with facts or themselves \citep{JiLee2023a}, or to be highly sensitive to spurious variations in their inputs/prompts \citep{SclarChoi2023}. 

Many machine learning models also suffer from their own complexity: consisting of millions, billions, or even trillions of parameters, they are black-boxes, opaque to human understanding. However, in order to reliably use machine learning models and AI systems based on such models, human users must be able to assess their limitations and deficiencies, and to understand the decisions that such systems make and why (codified, for example, as the right “to obtain an explanation of the decision reached” in the legal framework of the General Data Protection Regulation of the European Union; \citealp[][Recital 71]{GDPR2016}). Research in Explainable AI (XAI) addresses this need, and recent years have seen an explosion of explainability methods that aim to make the internal knowledge and reasoning of AI systems transparent and explicit, and thus interpretable and accessible to users. Explainability of model predictions is thus seen as a solution, and it is assumed that they enable users to construct functional ‘mental models’ \citep{Norman1983} of AI systems, i.e., models that closely correspond to the actual capabilities of the systems. 

Whether this is the case is an active research question and there is evidence that explainability comes with new challenges.
Important questions in XAI are what actually makes a good explanation, which criteria it needs to satisfy, and how the quality of explanations can be measured  \citep{AlshomaryLange2024}. Furthermore, recent perspectives emphasize that explanations should be social \citep{Miller2019} and constructed interactively, taking into account the user's explanation needs \citep{RohlfingCimiano2021.TCDS}. \citet{jacovi-goldberg-2020-towards} argue that evaluations of explanations should carefully distinguish plausibility (does it seem plausible to users) and faithfulness (does it reflect the model's internal reasoning) and that non-faithful, but plausible, explanations can be dangerous in that they let users construct faulty, and eventually dysfunctional, mental models that can lead to unwarranted trust \citep{JacoviMarasovic2021}.

\begin{figure}
    \centering
    \includegraphics[width=\columnwidth]{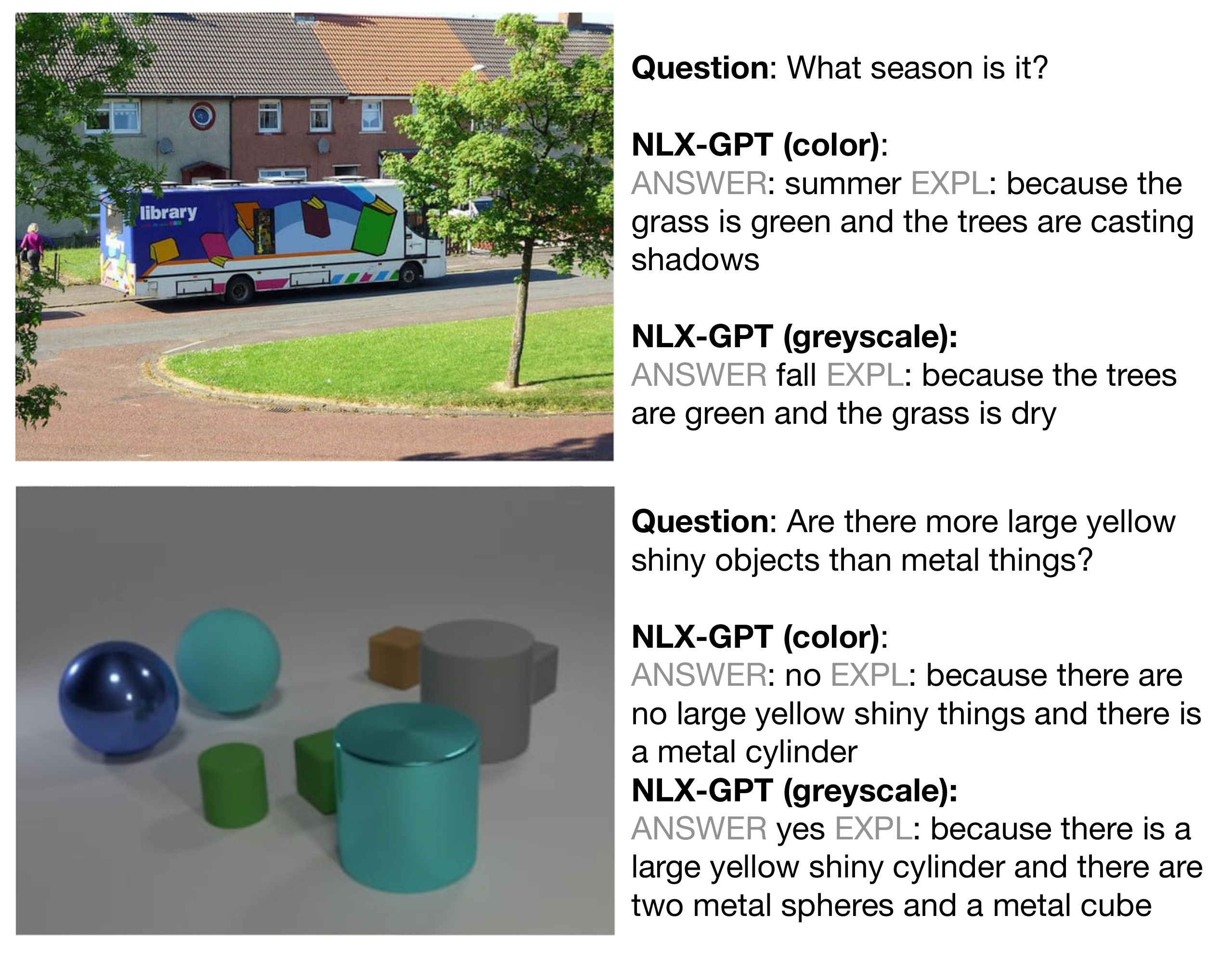}
    \caption{Items from our study: Answers and explanations generated with NLX-GPT for color/grayscale images in VQA-X (top) and CLEVR-X (bottom). Explanations in the grayscale condition refer to colors that were not available in the system inputs (\emph{green, yellow}).}
    \label{fig:dataset-example-images}
\end{figure}

In this paper, we investigate the effects of providing natural language explanations on users' mental models of an AI system in terms of its capabilities, and whether these explanations allow them to diagnose system limitations. We present the results of a study in the visual question answering and explanation (VQA/X) domain, artificially inducing a simple limitation by providing two VQA/X systems with images stripped of color information, i.e., in grayscale (see Figure~\ref{fig:dataset-example-images}). Participants, unaware of the manipulation, see the unmanipulated full color image, the question, the system's answer, and its explanation for the answer, and have to judge various system capabilities (including its ability to recognize colors) and its competence. This visual domain does not require participants to understand the internal processes of the system but should still enable them to estimate what it can and cannot do. The comparison of judgments to responses to non-manipulated system input and judgments of responses without explanations sheds light on participants' difficulties in using (natural language) XAI explanations to build accurate mental models, even for such a simple case. This raises the question of how effective explanations can be in real-world applications of XAI technology that involve more complex reasoning and problems.

\section{Background}
\label{sec:background}

Our work is related to previous studies that have examined whether explanations enhance users' trust in AI systems.
\citet{Kunkel2019-bf}, for example, compared trust in personal (human) versus impersonal (recommender system) recommendation sources and examined the impact of explanation quality on trust. Their results showed that users rated human explanations higher than system-generated ones and that the quality of explanations significantly influenced trust in the recommendation source.
\citet{Bansal2021} investigated whether explanations help humans anticipate when an AI system is potentially incorrect.
They used scenarios where an AI system helps participants to solve a task (text classification or question answering), providing visual explanations (highlighted words) under certain conditions.
Their findings revealed that explanations increased the likelihood of the participants to accept the AI system's recommendations, irrespective of their accuracy. Thus, rather than fostering appropriate reliance on AI systems, explanations tended to foster blind trust.
Similarly, \citep{Kim2021HIVEET} conducted a large-scale user study for visual explanations, showing that these do not allow users to distinguish correct from incorrect predictions.
\citet{dhuliawala-etal-2023-diachronic} investigated how users develop and regain trust in AI systems in human--AI collaborations. 
They found that NLP systems that confidently make incorrect predictions harm user trust, and that even a few incorrect instances can damage trust, with slow recovery. 
While these studies evaluate the influence of system explanations on users' trust in the system's output (a proxy for its perceived competence), they do not investigate users' understanding of the systems' reasoning processes and capabilities. In our study, we specifically address this issue and investigate the users' mental model of the systems' capabilities and limitations.

While the studies above found that nonverbal explanations can be misleading to users, natural language explanations are assumed to be more transparent or less difficult to interpret \citep{park2018multimodal, clevrx-Salewski2022}. Verbal explanations also offer the advantage that they can be collected from humans, which has led to the development of explanation benchmarks, particularly in multimodal domains \citep{Kayser_2021_ICCV, clevrx-Salewski2022}. Thus, the dominant approach to verbal explanation generation currently is to leverage human explanations during model training \citep{park2018multimodal, wu-mooney-2019-faithful, Kayser_2021_ICCV, plüster2023harnessing, sammani2023uninlx}. While \citet{lyu2024towards} discuss potential faithfulness issues related to supervising explanation generation with human explanations, we are not aware of work that explicitly tests these supervised models in a user-centered setting similar to ours.

\section{Approach}
\label{sec:approach}

\begin{figure*}
    \centering
    \includegraphics[width=1\textwidth]{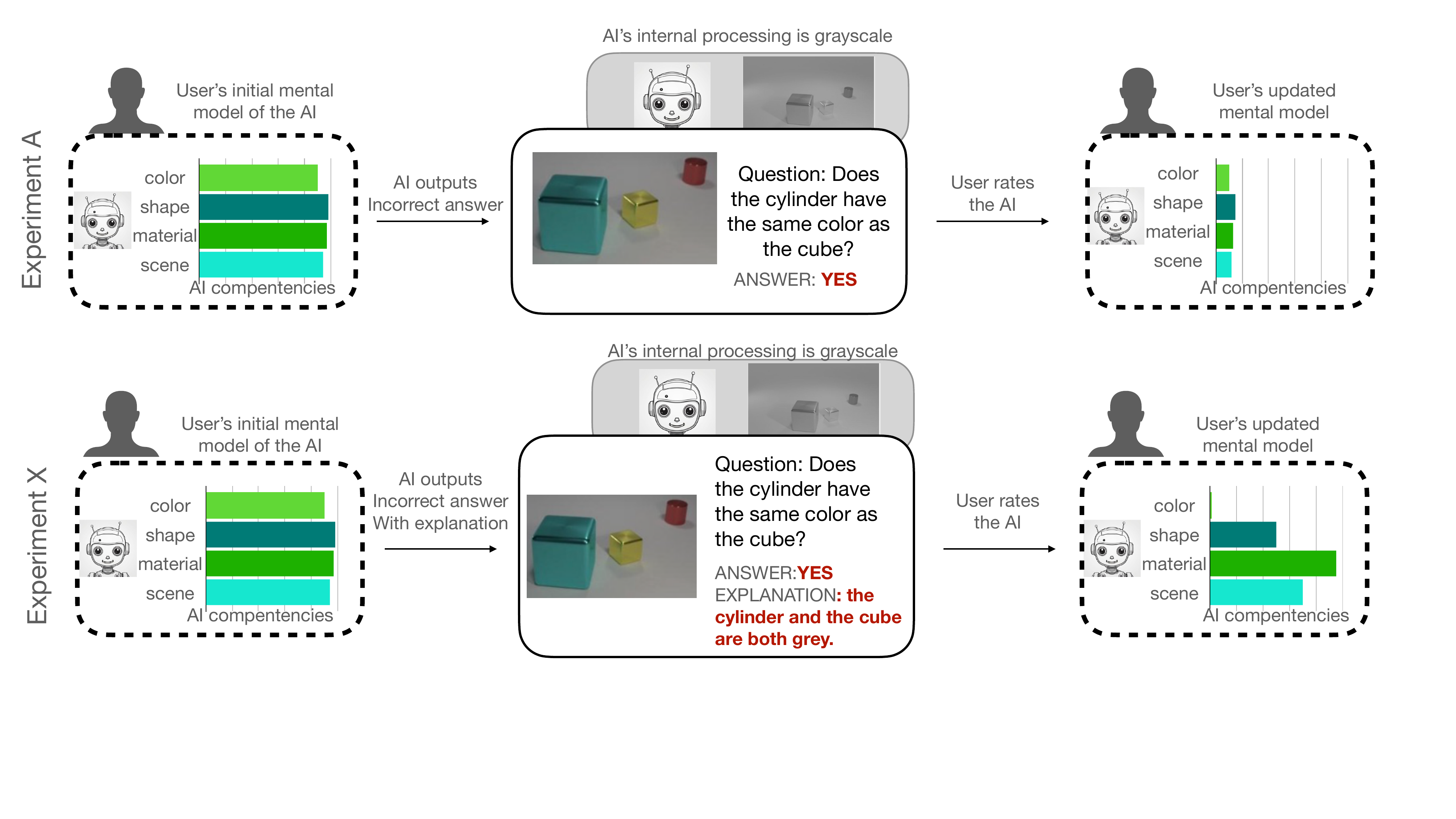}
    \vspace{-1.5cm}
    \caption{Illustration of our experimental design and hypotheses. In Exp.A, we do not expect users to spot the system defect (no color recognition due to grayscale input) since only answers are provided. In Exp.X, the system provides explanations which should help users in building a better mental model.}
    \label{fig:design}
\end{figure*}

We conduct a study to investigate how users of an AI system perceive its limitations when it encounters tasks that it cannot perform perfectly.
We aim to investigate whether providing explanations alongside model responses helps users build an appropriate mental model of the AI system's capabilities and limitations. In the following, we detail the rationale of our approach (Section~\ref{subsec:rationale}--\ref{subsec:grayscale}) and the hypotheses of our experiments (Section~\ref{subsec:experiments}).

\subsection{How Can We Test Users' Mental Models?}
\label{subsec:rationale}

A key challenge for our study is that many modern AI systems are used in complex tasks that involve many interdependent capabilities simultaneously.
This makes it difficult to isolate specific systems' capabilities and to establish or control which limitations they have, even for the developers of a system.
Indeed, common evaluation protocols in NLP mostly report overall system performance according to holistic metrics (e.g., accuracies) and rarely involve a detailed assessment of specific errors or capabilities \citep[cf.][]{van-miltenburg-etal-2021-underreporting}.
However, to assess whether users' perception of a system is accurate, we need to have as much control as possible over its capabilities and limitations.

To address this challenge, our study adopts an experimental setting that simplifies some aspects of testing XAI in complex tasks.
First, we focus on two well-studied VQA tasks, including a synthetic VQA task in which system capabilities are relatively easy to distinguish (Section~\ref{subsec:vqatask}).
Second, to gain at least some control over the VQA systems' limitations, we systematically manipulate their inputs in the dimension of color (Section~\ref{subsec:grayscale}).
Third, we design a questionnaire for users to judge specific aspects of the system's capabilities. This allows us to measure whether users can diagnose which capabilities of the system have been perturbed through our explicit input manipulations (Section~\ref{subsec:experiments}). 
The design of our study is summarized in Figure~\ref{fig:design} and will be explained in detail below.

\subsection{VQA Task and Abilities}
\label{subsec:vqatask}

We employ a visual question answering and explanation task: the input to the AI system is an image and a question in natural language, and its task is to generate an answer and a natural language explanation that justifies the answer. 
We select a visual question-answering setting as it is a rather simple task for humans and, at the same time, a task that involves distinguishable semantic-visual reasoning capabilities. This is important for our setting since we want to test whether users can differentiate specific system capabilities, based on generated explanations.
Thus, inspired by \citeauthor{clevrx-Salewski2022}'s (\citeyear{clevrx-Salewski2022}) CLEVR-X benchmark for explainable VQA, we assume that these capabilities involve the abilities to process objects' (i) \textbf{color}, (ii) \textbf{shape}, (iii) \textbf{material}, and (iv) \textbf{scene} composition (e.g., spatial relations, relative size). In our study participants are asked to rate the AI system's capabilities along these four dimensions, next to other, more general criteria for competence and fluency (see Figures~\ref{fig:screenshot_expA_vqax_trialrun} and \ref{fig:screenshot_expX_cleverx_trialrun} in Appendix~\ref{sec:appendix_online_experiment}).
In the CLEVR-X benchmark, these dimensions are given by construction: the visual scenes are synthetically generated and composed of objects defined by attributes for color, material, and shape. The corresponding questions explicitly relate to one or multiple of these dimensions. In real-world image benchmarks, such as VQA-X \citep{park2018multimodal}, these abilities are often more implicit, but still highly relevant (see examples in Figure~\ref{fig:dataset-example-images}). 
We run our study on items from both benchmarks.

\subsection{Color vs. Grayscale Input}
\label{subsec:grayscale}

Our goal is to investigate whether explanations help users in diagnosing system limitations. To introduce these limitations in a controlled way, we manipulate the input of the VQA systems. Out of the four VQA capabilities explained above (color, shape, material, and scene), the color dimension lends itself to straightforward manipulation. During inference, systems either receive the image (i) in full color or (ii) in grayscale. This induced limitation resembles a situation where a multimodal AI model was trained on colored images but, at run-time, a camera/visual sensor is broken such that model inputs are perturbed. To ensure that this manipulation induces an incorrect model response, we only include items correctly answered with the full color image input \emph{but} incorrectly answered with the grayscale image input. This item selection accounts for the fact that VQA models can be assumed to have further limitations that we cannot explicitly control for and exclude items (i) where the VQA does not generate the correct ground-truth answer for the colored image, and (ii) where the VQA generates the correct answer for the grayscale image. This gives us a clean set of items where the limitations of the AI system can be attributed to a particular error source.
The participants in our study were unaware of the underlying color--grayscale manipulation: they saw images in color, along with the models' answers and explanations. Our goal was to determine whether participants were able perceive the limitations of the model, i.e., whether they could identify the system's lack of color recognition ability. See Figure~\ref{fig:design} for an illustration of this set-up.

\subsection{Experiments A and X}
\label{subsec:experiments}

To investigate the effect of providing generated explanations alongside the system answers, we conduct two separate studies: In Experiment~X, participants were shown both the answer and its explanation, whereas in Experiment~A participants were shown only the answer without an explanation.
In both studies, we ask participants to rate each item for the system's capabilities (color, shape, material, scene), the overall system competence, answer correctness, the consistency of answer/explanation, the consistency of explanation/image, and the explanation's fluency.

Importantly, participants in both Experiments A and X received mixed sets of items from all systems, data sets, and color conditions, and we collected judgments for each item. In this way, we wanted to prevent them from becoming “conditioned” to a particular setting, i.e., getting used to certain ways of answering or explaining and becoming overly sensitive to changes in patterns.

If explanations lead users to build more appropriate mental models, participants should, generally speaking, be able to differentiate items where systems processed grayscale vs. full color images.
We approached this broad expectation with five hypotheses specific to our set-up (see Table~\ref{tab:hypotheses-results} for a brief summary). First, hypotheses H1\textsubscript{A} and H1\textsubscript{X} relate to the differences in competence scores between color and grayscale conditions. Here, we expect that explanations help participants to differentiate between different system capabilities. 

\begin{description}[noitemsep]
  
    \item[\textbf{H1\textsubscript{A}}]
    In Exp.A, competence and all capability scores are lower in the grayscale condition than in the color condition.
    
    \item[\textbf{H1\textsubscript{X}}]
    In Exp.X, competence and color capability scores are lower in the grayscale condition than in the color condition, but other capability scores are more stable.

\end{description}

   Hypotheses H2\textsubscript{A} and H2\textsubscript{X} are concerned with the comparison between individual competence scores in the grayscale condition. Again, explanations should help users to identify system deficiencies.

\begin{description}[noitemsep]
    
    \item[\textbf{H2\textsubscript{A}}]
    In the grayscale condition of Exp.A, participants give similar scores for all capabilities.
    
    \item[\textbf{H2\textsubscript{X}}]
    In the grayscale condition of Exp.X, participants rate the color capability lower relative to the other capabilities.

\end{description}

    Hypothesis H3\textsubscript{A/X} pertains to the comparison of competence scores between Exp.A and X. If explanations make defects in color processing transparent, grayscale inputs should specifically affect scores for this dimension. 

\begin{description}

    \item[\textbf{H3\textsubscript{A/X}}]
    In Exp.X the overall competence is rated higher than in Exp.A. In Exp.X, color competence is rated lower or the same as in Exp.A. 

\end{description}

\section{Experimental Setup}
\label{sec:experimental-setup}

\subsection{Data}
\label{sec:data}

We use two datasets in our study: VQA-X \citep{park2018multimodal} and CLEVR-X \citep{clevrx-Salewski2022}. 
VQA-X is extensively utilized in Visual Question Answering (VQA) tasks, as an extension of the well-established Visual Question Answering v1 \citep{VQA} and v2 \citep{balanced_vqa_v2} datasets.
The images within VQA-X originate from MSCOCO \citep{lin2015microsoft}, and the questions are open-ended (see Figure~\ref{fig:dataset-example-images}, top).
The style of the ground-truth explanations in VQA-X varies widely, ranging from simple image descriptions to detailed reasoning \citep{clevrx-Salewski2022}.

CLEVR-X expands the synthetic dataset CLEVR \citep{johnson2016clevr}, incorporating synthetic natural language explanations.
Each image in the CLEVR dataset depicts three to ten objects, each possessing distinct properties including size, color, material, and shape (see Figure~\ref{fig:dataset-example-images}, bottom).
For each image--question pair in the CLEVR dataset, CLEVR-X contains multiple structured textual explanations.
These explanations are constructed from the underlying scene graph, ensuring their accuracy without necessitating additional prior knowledge.

\subsection{Models}
\label{sec:models}

For each dataset, we used two vision and language models: (i) NLX-GPT \citep{Sammani2022-ta} and PJ-X \citep{park2018multimodal} for VQA-X, and (ii) NLX-GPT and Uni-NLX \citep{sammani2023uninlx} for CLEVR-X\footnote{%
    We tried to obtain model outputs from other explainable VQA-X models such as, e.g., OFA-X \citep{plüster2023harnessing}, FME \citep{wu-mooney-2019-faithful}, or e-UG \citep{Kayser_2021_ICCV}, but encountered significant reproducibility issues: code was unavailable or not running, authors were unavailable to provide model outputs, etc. \label{fn:model-outputs-reproducibility}
}. We did not use vanilla generative AI systems (such as ChatGPT) in this study, as we wanted to investigate models that were specifically constructed to provide explanations alongside their outputs.

NLX-GPT is an encoder--decoder model, which combines CLIP \citep{RadfordKim2021CLIP} as the visual encoder with a distilled GPT-2 model \citep{Radford2019}. 
Importantly, this model jointly predicts answers and explanations, i.e., it generates a single response string of the form “the answer is <answer> because <explanation>”, given a question and image. 
For VQA-X, we use the model from \citet{Sammani2022-ta}, which is pre-trained on image-caption pairs and fine-tuned on the VQA-X data. For CLEVR-X, we use the published pre-trained weights and fine-tune the model on this dataset.
Uni-NLX relies on the same architecture as NLX-GPT, but the model is trained on various datasets for natural language explanations (including VQA-X), to leverage shared information across diverse tasks and increase flexibility in both answers and explanations. We take the trained model from \citet{sammani2023uninlx} and fine-tune it on CLEVR-X.
While NLX-GPT and Uni-NLX generate answers and explanations simultaneously, the PJ-X model takes a two-step approach.
It first predicts the answer with an answering model and, subsequently, generates visual and textual explanations based on the question, image, and answer\footnote{%
    We could not replicate \citeauthor{clevrx-Salewski2022}'s (\citeyear{clevrx-Salewski2022}) PJ-X results on CLEVR-X, and the authors could not provide model outputs. Therefore, we only report PJ-X on VQA-X.\label{fn:PJ-X-reproducibility}}.

For each model, we utilize the recommended model weights and fine-tune them on the two datasets. During fine-tuning, we supply each model with the original, i.e., full color images along with the questions, answers, and explanations for both datasets.
During inference, images are presented in color alongside the question, or in grayscale.

\subsection{User Study}

We conducted the study online, using \href{https://www.prolific.co/}{Prolific}, and obtained ratings from 160 participants (80 each in Exp.A and X) who were native English speakers with normal color vision (selected using Prolific's filters). 
In both experiments, we utilized identical experimental items, differing only in the presence or absence of explanations. 
All items consisted of instances where the model provided correct answers for colored images and incorrect answers for grayscale images.
We selected a total of 128 items, evenly distributed across the datasets and models, comprising 64 for each dataset and 32 for each model, equally split between 16 colored and 16 grayscale items (for NLX-GPT, a total of 64 items were selected, with 32 items from CLEVR-X and 32 items from VQA-X). 
The items were distributed over four experimental lists, with each participant evaluating 32 individual items.
We gathered 2560 judgments per experiment and 5120 overall.

We designed the evaluation as a rating task. We informed participants that we are assessing an AI system's ability to answer questions about images (and, for Exp.X, to generate explanations). The image, question, and answer for each item were presented at the top of the page, and, in Exp.X, the generated explanation was displayed below the answer.
Each item had several questions and statements for the participants to assess.
First, they were asked to evaluate the correctness of the answer.
In Exp.X, participants were further asked to assess whether the explanation was (i) consistent with the answer, (ii) consistent with the picture, and (iii) overall fluent.
Additionally, participants in both experiments were asked to judge whether they believed that the AI system correctly identifies (iv) shapes, (v) colors, and (vi) materials, as well as whether it (vii) understands the general scene in the image. Finally, (viii) participants judged the overall competence of the system.
Participants indicated their agreement on five-point Likert scales, ranging from 1 (‘strongly disagree’) to 5 (‘strongly agree’). For each criterion, we also offered the option of selecting “I don't know”.
Before providing ratings, participants received instructions and viewed an example item illustrating the evaluation criteria. They were paid at a rate of £9.00 per hour. See Appendix~\ref{sec:appendix_additional_results} for example trials of the experiment.

\section{Results}
\label{sec:results}

We organize the discussion of results based on the hypotheses outlined in Section~\ref{sec:approach}.
Since we ask whether explanations help participants determine that the systems could not recognize color, the following discussion concentrates on the grayscale condition and the differences between the grayscale and color conditions (see Appendix~\ref{sec:appendix_additional_results} for detailed results of the color condition).

\begin{figure*}[ht]
    \centering
    \begin{subfigure}{0.24\textwidth}
        \centering
        \includegraphics[width=\linewidth]{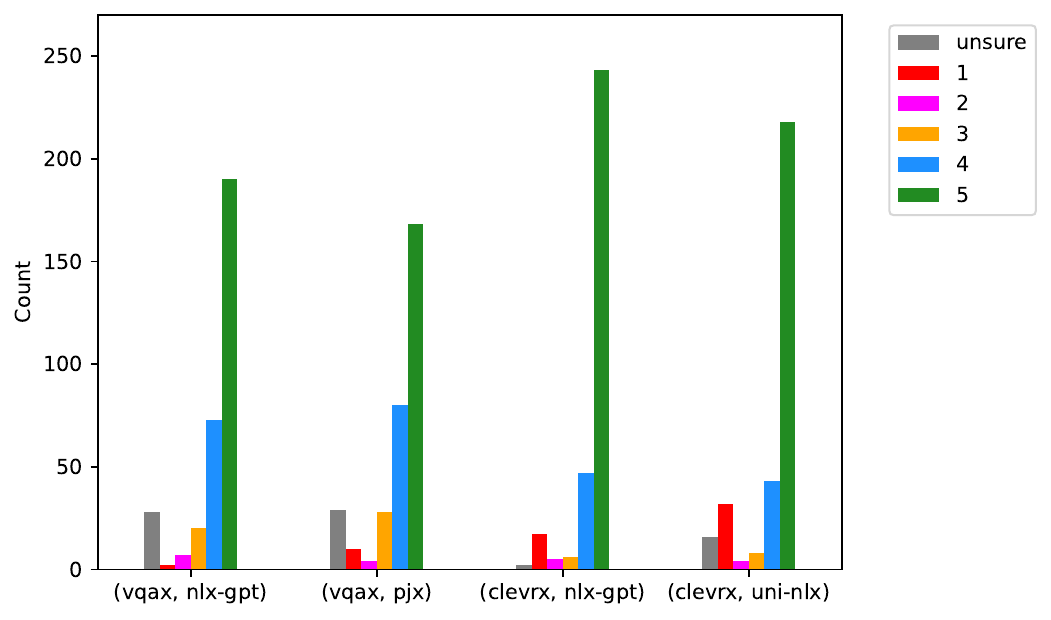}
        \caption{Exp.A -- colored images}
    \end{subfigure}
    \hfill
    \begin{subfigure}{0.24\textwidth}
        \centering
        \includegraphics[width=\linewidth]{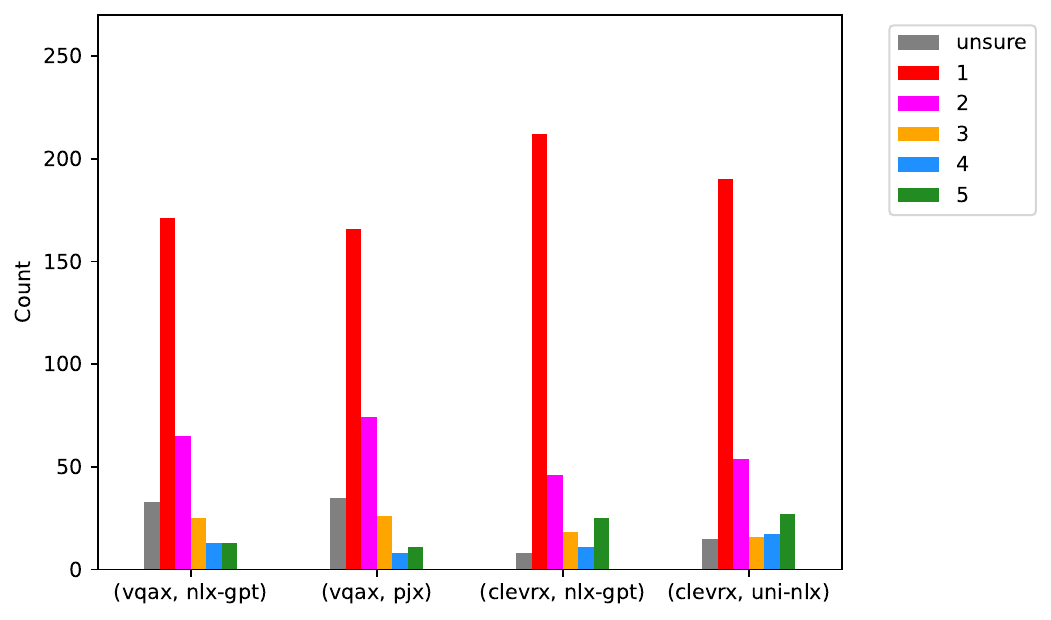}
        \caption{Exp.A -- grayscale images}
    \end{subfigure}
    \hfill
    \begin{subfigure}{0.24\textwidth}
        \centering
        \includegraphics[width=\linewidth]{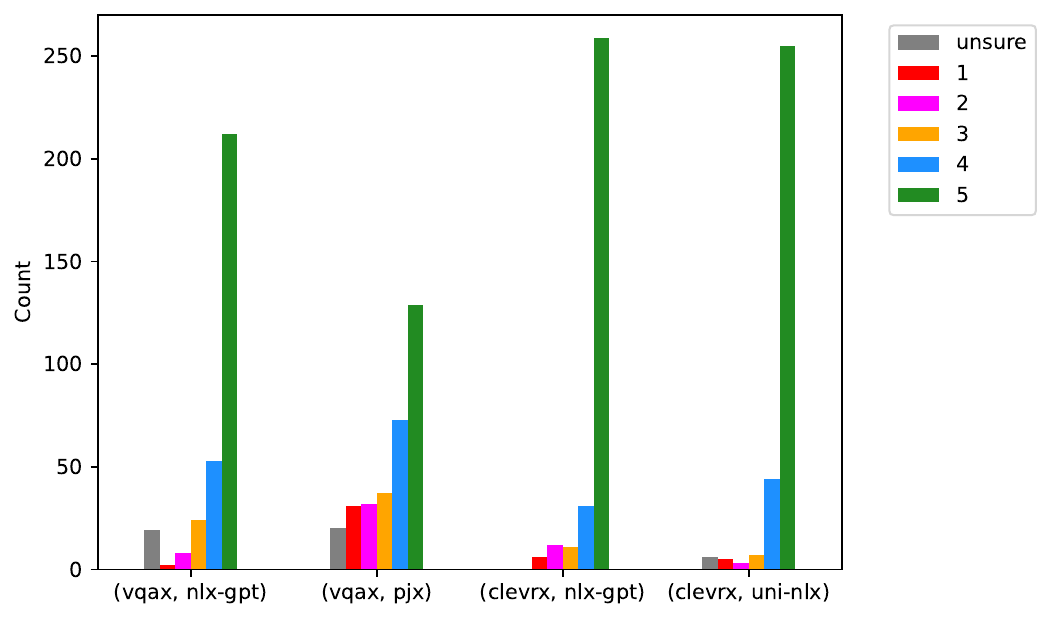}
        \caption{Exp.X -- colored images}
    \end{subfigure}
    \hfill
    \begin{subfigure}{0.24\textwidth}
        \centering
        \includegraphics[width=\linewidth]{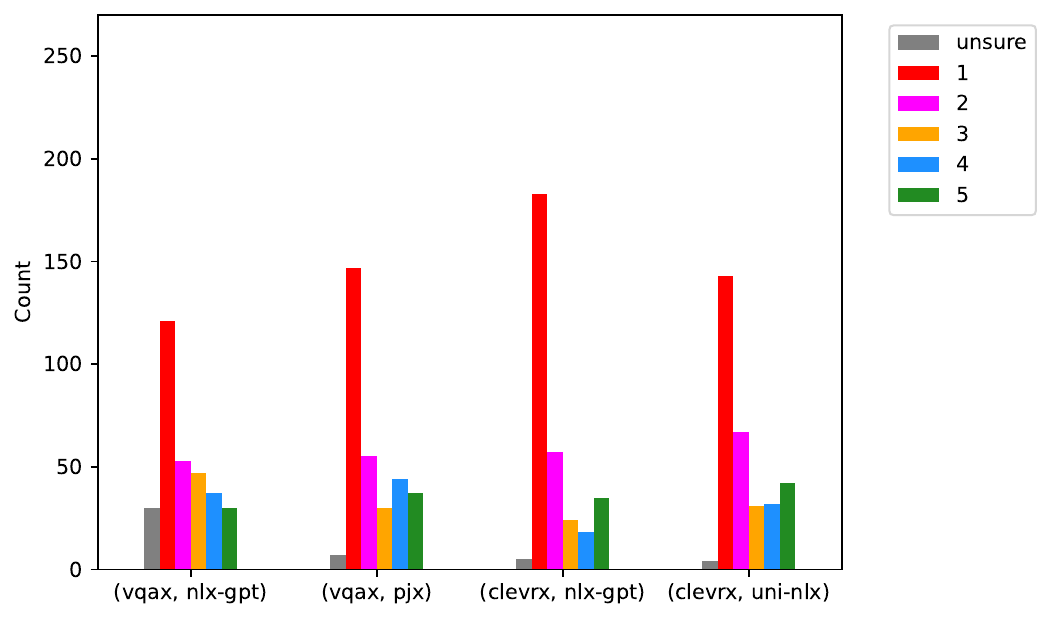}
        \caption{Exp.X -- grayscale images}
    \end{subfigure}
    
    \caption{Human ratings on the evaluation criterion “Ability of the AI to \textbf{recognize colors}”. Participants indicated their judgment on a scale from 1 (strongly disagree; here in red) to 5 (strongly agree; here in green).}  
    \label{fig:plots_colorabil}
\end{figure*}

All systems received high ratings in all competency and capability dimensions when tested in the color condition of Exp.A and X, on both datasets (see Table~\ref{tab:ExpAX_results_color_cond_allcriteria} in Appendix~\ref{sec:appendix_additional_results}). These ratings decreased in very similar ways in the grayscale condition. Therefore, we were able to use all items from all systems to test our hypotheses, generalizing over minor system differences. 
We discuss differences between datasets and models in Appendix~\ref{sec:appendix_additional_results}, since these were not essential for testing our hypotheses. See Table~\ref{tab:hypotheses-results} for summaries of hypotheses and results.

\subsection[Hypotheses H1-A and H1-X]{Hypotheses H1\textsubscript{A} and H1\textsubscript{X}}

Hypotheses H1\textsubscript{A} and H1\textsubscript{X} state our expectations on distinctions between the grayscale and color conditions in Exp.A and X, respectively.
Figure~\ref{fig:plots_colorabil} shows the distribution of participant ratings for the AI system's ability to recognize colors, for the grayscale and color conditions in both experiments (see Figures~\ref{fig:plots_genscene}, \ref{fig:plots_competency}, \ref{fig:plots_shapes}, and \ref{fig:plots_materials} in Appendix~\ref{sec:appendix_additional_results} for results on the other capabilities).
In Exp.A and X, there is a consistent trend of better assessments when systems have been seen the color images compared to grayscale images, across different systems, datasets, and all capabilities. Most users rate the color capability with the highest rating in the color condition (Figure~\ref{fig:plots_colorabil}a/c) and with the lowest rating in the grayscale condition (Figure~\ref{fig:plots_colorabil}b/d). The same holds for all other capabilities and competency (Figures~\ref{fig:plots_genscene}, \ref{fig:plots_competency}, \ref{fig:plots_shapes}, and \ref{fig:plots_materials}).
This confirms hypothesis H1\textsubscript{A}, i.e., ratings for all capabilities decrease when the system does not see color. However, this does not support H1\textsubscript{X}, as we expected that only overall competence and capability to recognize colors would be rated lower in the grayscale condition when explanations were given, and not all capabilities. This suggests that the AI's explanations did not help users diagnose the system's limitation in the grayscale condition, as all capability dimensions are similarly affected in Exp.X.

\begin{table*}[!htbp]
\small
\centering
\begin{tabularx}{\textwidth}{lllYYYYYYYYYY}
\toprule
    & &     & \multicolumn{2}{c}{Colors} & \multicolumn{2}{c}{Shapes} & \multicolumn{2}{c}{Materials} & \multicolumn{2}{c}{General Scene} & \multicolumn{2}{c}{Competency} \\
     \cmidrule(rl){4-5} \cmidrule(rl){6-7} \cmidrule(rl){8-9} \cmidrule(rl){10-11} \cmidrule(rl){12-13}
Experiment & Dataset     & Model     & med &  mean & med &  mean &    med &  mean &    med &  mean &    med &  mean \\
\midrule
\textbf{Exp.A} &CLEVR-X  & NLX-GPT &    1.0 &  1.69 &    1.0 &  2.08 &       1.0 &  1.94 &       1.5 &  1.97 &       1.0 &  1.68 \\
&     & Uni-NLX &    1.0 &  1.84 &    2.0 &  2.31 &       1.0 &  2.11 &       2.0 &  2.16 &       1.0 &  1.91 \\
& VQA-X & NLX-GPT &    1.0 &  1.73 &    2.0 &  2.23 &       1.0 &  1.71 &       1.0 &  1.87 &       1.0 &  1.64 \\
&     & PJ-X &    1.0 &  1.71 &    2.0 &  2.08 &       1.0 &  1.74 &       1.0 &  1.83 &       1.0 &  1.60 \\
\midrule
\textbf{Exp.X} & CLEVR-X  & NLX-GPT &    1.0 &  1.93 &    3.0 &  2.95 &       2.0 &  2.62 &       2.5 & 2.61 &       2.0 &  2.13 \\
&      & Uni-NLX &    2.0 &  2.27 &    3.0 &  2.89 &       3.0 &  2.82 &       2.0 &  2.61 &       2.0 &  2.21 \\
& VQA-X & NLX-GPT &    2.0 &  2.36 &    3.0 &  2.70 &       2.0 &  2.32 &       2.0 &  2.29 &       2.0 &  1.96 \\
&      & PJ-X &    2.0 &  2.25 &    2.0 &  2.53 &       2.0 &  2.32 &       2.0 &  2.23 &       2.0 &  1.88 \\
\bottomrule
\end{tabularx}
\caption{Human ratings on system capabilities for the \textbf{grayscale condition} of Exp.A (no explanations) and Exp.X (with explanations), as median and mean scores across raters.}
\label{tab:ExpAX_results_grayscale_cond_allcriteria}
\end{table*}

\subsection[Hypotheses H2-A and H2-X]{Hypotheses H2\textsubscript{A} and H2\textsubscript{X}}

Hypotheses H2\textsubscript{A} and H2\textsubscript{X} state our expectations for the grayscale condition.
Table~\ref{tab:ExpAX_results_grayscale_cond_allcriteria} presents the human evaluation results in Exp.A and X. 
Starting with Exp.A, Table~\ref{tab:ExpAX_results_grayscale_cond_allcriteria} shows that all evaluation criteria in the grayscale condition receive relatively low scores. Interestingly, the manipulated capability, i.e., to recognize colors, does have slightly worse ratings than the other criteria (for most models and datasets). 
This outcome does not align with our expectation (H2\textsubscript{A}) as participants in Exp.A solely viewed the answers without access to explanations, making it difficult to discern which specific ability or (limitation) influenced the model's answer. 
Results from  Mann-Whitney U tests (see Table~\ref{tab:grayscale_mann-whitney-u-test} in Appendix~\ref{app:sigtest}) show significant differences between the ability to recognize colors and the ability to recognize other criteria for Exp.A (except for the models' overall competence), contradicting hypothesis (H2\textsubscript{A}). This suggests that users in Exp.A were able to interpret incorrect system answers more than we expected.
\begin{table*}
\centering
\small
\begin{tabularx}{\textwidth}{lXc}
\toprule
H1\textsubscript{A}
    & competence and all capabilities rated lower in grayscale cond. than in color cond. in Exp.A  
    & \CheckmarkBold \\
H1\textsubscript{X}
    & competence and color capability rated lower in grayscale cond. than in color cond. in Exp.X 
    & \XSolidBrush\\
H2\textsubscript{A} 
    & similar ratings for color compared to other capabilities, in grayscale cond. in Exp.A  
    & \XSolidBrush\\
H2\textsubscript{X} 
    & lower ratings for color compared to other capabilities, in grayscale cond. in Exp.X 
    & (\Checkmark)\\
H3\textsubscript{A/X} 
    & competence rated higher for grayscale cond. in Exp.X than in Exp.A, color rated lower 
    & (\Checkmark/\XSolidBrush)\\
\bottomrule
\end{tabularx}
\caption{Overview of the validity of the hypotheses formulated in Section~\ref{sec:approach}.}
\label{tab:hypotheses-results}
\end{table*}
For Exp.X, the results in Table~\ref{tab:ExpAX_results_grayscale_cond_allcriteria} suggest a very similar trend to Exp.A: the ability to recognize colors is rated slightly lower than the other capabilities.
The Mann-Whitney U tests for Exp.X (reported in the lower part of Table~\ref{tab:grayscale_mann-whitney-u-test} in Appendix~\ref{app:sigtest} ), again confirms significant differences between the perceived ability to recognize colors and the other abilities (except the systems' overall competence). 
Looking at Exp.X in isolation, these results seem to speak in favor of our hypothesis H2\textsubscript{X}: users were indeed able to diagnose the system defect, at least to some extent. However, in light of our findings on H2\textsubscript{A}, these results have to be interpreted with care: even without model explanations, users rated the color capability lower than others.
This trend is a bit stronger in Exp.X but, overall, the differences between perceived capabilities are still rather small. The strongest expected trend in favor of H2\textsubscript{X} can be found for NLX-GPT on the CLEVR-X data: here, the median if the color rating is $1.0$ and $3.0$ or $2.0$ for the other capabilities. For the other combinations of models and datasets in Exp.X, there is no clear difference in the median ratings for the perceived capabilities.
We conclude that there is weak evidence in favor of H2\textsubscript{X}, as explanations do not substantially improve users' assessments of system capabilities.

\subsection[Hypothesis H3-A/X]{Hypothesis H3\textsubscript{A/X}}

Hypothesis H3\textsubscript{A/X} states our expectations regarding the differences between Exp.A and X for overall competency and color recognition ability. 

Once again, consider Table~\ref{tab:ExpAX_results_grayscale_cond_allcriteria}.
As expected, in Exp.A, i.e., without explanations, the overall competency of the models was rated low (with median values of $1.0$ only). 
In Exp.X, although the values remain low at 2.0, there is a noticeable improvement relative to Exp.A. Thus, despite the answers being incorrect, the addition of the models' explanations enhances the perception of the models' overall competency. This could suggest that the explanations reveal other capabilities of the models, consistent with our hypothesis H3\textsubscript{A/X}. 
However, contrary to H3\textsubscript{A/X}, we also see a general increase in the ratings for the systems' color recognition ability in Exp.X compared to Exp.A. We expected that the explanations would make the color limitation explicit, which would result in color ability being rated worse or at least as poorly as in Exp.A. This also holds for \emph{all} other model capabilities: all capability ratings are comparatively higher in Exp.X than in Exp.A (even if lower than in the color condition). This observation is supported by the Mann-Whitney U tests  (see the upper part of Table~\ref{tab:grayscale_mann-whitney-u-test} in Appendix~\ref{app:sigtest}), which show significant differences between Exp.A and X for all evaluation criteria. This suggests that users rate all system capabilities significantly higher when explanations are provided.
From this we conclude that, instead of making systems' limitations more transparent, the explanations contribute to an overall more positive perception of the system, regardless of its capabilities. In other words, the AI system's explanations seem to create an illusion of the system's competence that does not correspond to its actual performance.

\subsection{Automatic Evaluation}

In the VQA-X domain, automatic measures for evaluating similarity or overlap with human ground-truth explanations are commonly used \citep[cf.][]{clevrx-Salewski2022,sammani2023uninlx}. To assess the construct validity of a representative automatic evaluation method, we compute BERTScores, measuring the similarity of ground truth explanations from both datasets to human evaluation scores. 
Table~\ref{tab:BERTScore-results} reports the results of the BERTscore metric, showing that they do not exhibit any notable differences between the grayscale and color conditions, which clearly contradicts the results of our human investigation. Thus, while user ratings between the grayscale and color condition are located on opposite ends on the Likert scale, BERTscores show marginal differences across the board. Yet, when comparing the two datasets, the BERTScores for the CLEVR-X dataset show improved values (in both the grayscale and color conditions), aligning with the human results from Exp.X (see Table~\ref{tab:ExpAX_results_grayscale_cond_allcriteria} and \ref{tab:ExpAX_results_color_cond_allcriteria} in Appendix~\ref{sec:appendix_additional_results}).

\begin{table}
\small
\begin{tabularx}{\columnwidth}{@{}XXYY@{}}
\toprule
    && \multicolumn{2}{c}{BERTScore} \\
\cmidrule{3-4}
    Dataset & Model   & color & grayscale \\
\midrule
    CLEVR-X & NLX-GPT & 0.76  & 0.74 \\
            & Uni-NLX & 0.75  & 0.74 \\
    VQA-X   & NLX-GPT & 0.72  & 0.72 \\
            & PJ-X    & 0.71  & 0.70 \\
\bottomrule
\end{tabularx}
\caption{BERTScores for explanations by condition.}
\label{tab:BERTScore-results}
\end{table}

\subsection{Summary}

Table~\ref{tab:hypotheses-results} provides an overview of the validity of our hypotheses. Generally, our results show that explanations do not have a desirable effect on users' assessment of the system's competency and capabilities. They do not help users construct a more accurate mental model of the system and its capabilities and limitations, but simply lead to more positive user assessment overall.
Our results are strikingly consistent across models and datasets. Even systems fine-tuned on the CLEVR-X benchmark, where explanations were designed to systematically mention the capabilities we assessed in our study (including color), do not address these limitations. Figure~\ref{fig:dataset-example-images} shows representative examples of why this might be the case: rather than avoiding color words or using incorrect colors, systems seem to be able to guess the correct color from the question or the general context (e.g., \emph{green} in the context of \emph{tree}). This behavior is well-known in multimodal language models but should be avoided in explanation tasks since it counteracts transparency and appropriate user assessment.

\section{Discussion of Implications}

It is still not well understood how XAI can bridge the gap between highly complex black-box models with largely opaque internal reasoning processes and users' intuitive understanding of these. Generally, our study provides evidence that explanations generated by state-of-the-art systems do not always lead to the expected effects of XAI and that explanations may even further obstruct AIs' reasoning processes and trick users into believing that the AI is more competent than it actually is. This result is particularly noteworthy in light of the fact that the manipulation employed in our study introduced an obvious error that should be easy to spot for users (defects in systems' color recognition).

\paragraph{XAI Models}

Our study underlines the great importance of prioritizing faithfulness over plausibility in explanation methods \citep{jacovi-goldberg-2020-towards}. With today's AI systems and LLMs, users face the challenging situation that these systems present fluent outputs projecting confidence and competence. Yet, this confidence may not be grounded in actual system capabilities and reliability \citep{GuoPleiss2017}. Our findings suggest that this also holds, to some extent, for state-of-the-art approaches to natural language explanation generation. Looking at the architecture of these models, this is by no means surprising. At least within the domain of VQA-X, which we focused on in this paper, explanation generation approaches largely follow common language modeling architectures and prioritize generating fluent, human-like outputs. Despite the fact that the importance of faithfulness in XAI has been recognized for some time and it continues to be a challenge \citep{lyu2024towards}.

\paragraph{Evaluation of XAI} 

Our study also highlights the importance of evaluating explanation methods in thorough, detailed, and user-centered ways \citep[cf.][]{lopes2022survey}. In the domain of VQA-X, automatic, benchmark-based evaluations still seem to be in focus and widely accepted in the community. All systems we tested in our study have been assessed mainly in automatic evaluations \citep[cf. ][]{park2018multimodal, Kayser_2021_ICCV, Sammani2022-ta, sammani2023uninlx}. This stands in stark contrast to research showing that XAI evaluations often have little construct validity, i.e., do not assess the intended properties of explanations \citep{DoshiVelez2017TowardsAR, VANDERWAA2021}. Our BERTscore-results lend further support to this argument.

\section{Conclusion}
\label{sec:conclusion}

This paper investigates the effects of providing natural language explanations on users' ability to construct accurate mental models of AI systems' capabilities, and whether these explanations allow them to diagnose system limitations. Results from two experiments show that natural language explanations generated by state-of-the-art VQA-X systems may actually hinder users from accurately reflecting capabilities and limitations of AI systems. Participants who received natural language explanations projected more competence onto the system and rated its limited capabilities higher than those who did not receive explanations.

\section*{Limitations} 

We identify the following limitations in our work:

The addition of further models and data sets might have provided additional insights into our experiments. Unfortunately, recently research on generating natural language explanations has not been very active. The best known approaches are models like PJ-X \citep{park2018multimodal} or e-UG \citep{Kayser_2021_ICCV}, which have older code bases with reproducibility issues. We have tried to include other models (see Section~\ref{sec:experimental-setup}, \cref{fn:model-outputs-reproducibility,fn:PJ-X-reproducibility}).

For the grayscale condition, we remove color information at the inference level for models trained on colored input. An alternative approach would be altering inputs during model training, possibly leading to deficiencies that are harder to identify for participants.
Similarly, other kinds of perturbations such as altering relative object sizes or scene layouts might affect different dimensions of perceived system capabilities than color recognition. Here, we focused on color, as this property is easier to control and less intertwined with other properties than, e.g., object size (which might also change how relative positions are described).

\section*{Ethics Statement}

Our study focuses on user-centered evaluation of XAI systems and on understanding whether these systems fulfill the promise of making black-box AI systems more transparent for users. Therefore, we believe that our study contributes to understanding and improving the social and ethical implications of recent work in NLP, and Language \& Vision. In our study, we collect ratings from Prolific users but, other than that, did not record any personal information on these users.

\section*{Acknowledgments}

The first author acknowledges financial support by the
project “SAIL: SustAInable Life-cycle of Intelligent Socio-Technical Systems" (Grant ID NW21-059A), an initiative of the Ministry of Culture and
Science of the State of North Rhine-Westphalia. We also acknowledge funding and support by the \href{https://www.dfg.de}{Deutsche Forschungsgemeinschaft (DFG)} (\href{https://gepris.dfg.de/gepris/projekt/438445824}{TRR 318/1 2021 – 438445824}) and Honda Research Institute Europe.


\bibliography{emnlp2024-bibliography}
\vfill\pagebreak

\appendix
\section{Appendix}
\label{sec:appendix}

\FloatBarrier

\subsection{Materials Availability Statement}
\label{app:materials-availablity}
We used the following public resources in our work:

\begin{itemize}
    \item Source code for NLX-GPT is available from GitHub at \\ \href{https://github.com/fawazsammani/nlxgpt}{https://github.com/fawazsammani/nlxgpt}
    \item Source code for Uni-NLX is available from GitHub at \\ \href{https://github.com/fawazsammani/uni-nlx/}{https://github.com/fawazsammani/uni-nlx/}
    \item Source code for PJ-X and VQA-X data is available from GitHub at \href{https://github.com/Seth-Park/MultimodalExplanations}{https://github.com/Seth-Park/MultimodalExplanations}
    \item COCO Images for VQA-X are available here: \href{https://cocodataset.org/}{https://cocodataset.org/}
    \item CLEVR-X data is available from GitHub at \href{https://github.com/ExplainableML/CLEVR-X}{https://github.com/ExplainableML/CLEVR-X}
    \item CLEVR images for \mbox{CLEVR-X} are available here: \\ \href{https://cs.stanford.edu/people/jcjohns/clevr/}{https://cs.stanford.edu/people/jcjohns/clevr/}
\end{itemize}

The source code and data from our human evaluation study can be found at either of the following
locations:

\begin{itemize}

    \item \href{https://doi.org/10.17605/OSF.IO/4KDB5}{https://doi.org/10.17605/OSF.IO/4KDB5}

    \item  \href{https://github.com/clause-bielefeld/IllusionOfCompetence-VQA-Explanations}{https://github.com/clause-bielefeld/ IllusionOfCompetence-VQA-Explanations.}
    
\end{itemize}

\subsection{Statistical Tests}
\label{app:sigtest}

Table~\ref{tab:grayscale_mann-whitney-u-test} shows the results of Mann-Whitney U tests in the grayscale condition. The upper half of the table reports the differences in user ratings of system capabilities (color, shape, material, scene) and overall competence between Exp.A and X, all differences are highly statistically significant. The lower half of the Table reports the differences in ratings with Exp.A and X. Table~\ref{tab:color_mann-whitney-u-test} reports the same tests for the color condition. Here, only the difference between overall competence is statistically significant between Exp.A and X while all system capabilities are rated similarly with or without explanations. This further supports our finding that explanations enhance user's perception of system competence, regardless of the correctness of system answers.

\begin{table}[ht]
\centering
\small
\begin{tabularx}{\columnwidth}{lcr}
\toprule
    Criterion & U-statistic & $p$-value \\
\midrule
    Colors        & 488421.0 & \textbf{$\mathbf{4.09\times10^{-15}}$} \\
    Shapes        & 460501.0 & \textbf{$\mathbf{5.81 \times 10^{-21}}$} \\
    Materials     & 428263.0 & \underline{\textbf{$\mathbf{3.06 \times 10^{-32}}$}} \\
    General Scene & 457629.0 & \textbf{$\mathbf{3.38 \times 10^{-22}}$} \\
    Competency    & 464419.5 & \textbf{$\mathbf{3.01 \times 10^{-21}}$ }\\
\midrule
    Color / Shape (Exp.A)      & 452212.0 & \textbf{$\mathbf{1.64 \times 10^{-15}}$} \\
    Color / Shape (Exp.X)      & 506384.0 & \textbf{$\mathbf{4.70 \times 10^{-21}}$} \\
    Color / Material (Exp.A)   & 510967.5 & \textbf{$\mathbf{6 \times 10^{-04}}$} \\
    Color / Material (Exp.X)   & 548762.5 & \textbf{$\mathbf{3.43 \times 10^{-11}}$} \\
    Color / Gen. Scene (Exp.A) & 486718.0 & \textbf{$\mathbf{1.70 \times 10^{-06}}$} \\
    Color / Gen. Scene (Exp.X) & 557231.0 & \textbf{$\mathbf{4.54 \times 10^{-09}}$}\\
    Color / Comp. (Exp.A)      & 538178.0 & 0.52 \\
    Color / Comp. (Exp.X)      & 640143.5 & 0.73 \\
\bottomrule
\end{tabularx}
\caption{Mann-Whitney U test results for the \textbf{grayscale conditions} of Experiments A and X. In the upper part of the table, we measure whether the ratings of one evaluation criterion (e.g., the ability to recognize \emph{colors}) of Exp.A differs significantly from the ratings of the same evaluation criterion from Exp.X. In the lower part of the table, we measure whether the ratings of the color criterion differ significantly from the ratings of the other evaluation criteria. $p$-values in bold indicate statistical significance ($p < 0.001$), the smallest $p$-value is underlined.} 
\label{tab:grayscale_mann-whitney-u-test}
\end{table}

\begin{table}[ht]
\centering
\small
\begin{tabularx}{\columnwidth}{lcr}
\toprule
    Criterion     &  U-statistic & $p$-value \\
\midrule
    Colors        & 627628.0     & 0.77510 \\
    Shapes        & 632776.5     & 0.49522 \\
    Materials     & 606350.0     & 0.17573 \\
    General Scene & 647675.0     & 0.06266 \\
    Competency    & 678234.5     & \underline{\textbf{0.00003}} \\
\midrule
    Colors / Shapes (Exp.A)     & 594055.5 & 0.23511 \\
    Colors / Shapes (Exp.X)     & 706324.0 & 0.14946 \\ 
    Colors / Materials (Exp.A)  & 626865.0 & \textbf{0.00012}  \\
    Colors / Materials (Exp.X)  & 717614.5 & \textbf{0.02390} \\
    Colors / Gen. Scene (Exp.A) & 569399.0 & 0.84294 \\
    Colors / Gen. Scene (Exp.X) & 710226.5 & 0.08423 \\
    Colors / Competency (Exp.A) & 572890.5 & 0.61815 \\
    Colors / Competency (Exp.X) & 746006.5 & \underline{\textbf{0.00002}} \\
\bottomrule
\end{tabularx}
\caption{Mann-Whitney U test results for the \textbf{color conditions} of Experiments A and X. In the upper part of the table, we measure whether the ratings of one evaluation criterion (e.g. the ability to recognize \emph{colors}) of Exp.A differs significantly from the ratings of the same evaluation criterion from Exp.X. In the lower part of the table, we measure whether the ratings of the color criterion differs significantly from the ratings of the other evaluation criteria. $p$-Values in bold indicate significance ($p < 0.05$), the smallest $p$-values are underlined.} 
\label{tab:color_mann-whitney-u-test}
\end{table}

\subsection{Additional Results}
\label{sec:appendix_additional_results}

\paragraph{Answer Correctness} 
First, recall that we only included cases where the models generated incorrect answers for grayscale images and correct answers for full-color images, according to ground-truth answers in the datasets. Table~\ref{tab:answer_correct_ExpAandB} displays frequency distributions of correctness ratings in our user study: ‘no’ ratings predominated in the grayscale condition, whereas ‘yes’ ratings were more prevalent in the color condition across both datasets. We also conducted a chi-squared test of independence on this evaluation criterion ($\chi^2 = 2.3617, df=2, p = 0.67$), finding no statistically significant difference between Exp.A and X regarding the evaluation of the answers' correctness. These results replicate and confirm the correctness of ground-truth answers in VQA-X and CLEVR-X.

\begin{table*}[!htb]
\centering
\small
\begin{tabularx}{\textwidth}{lYYYYYY}
\toprule
& \multicolumn{3}{c}{Exp.A} & \multicolumn{3}{c}{Exp.X} \\
\cmidrule(lr){2-4} \cmidrule(lr){5-7}
    Condition  & no & unsure & yes & no & unsure & yes \\
\midrule
    grayscale  & 1129 & 51 &   99 & 1157 & 36 &   86\\
    color      &   82 & 67 & 1131 &   59 & 48 & 1172\\
\bottomrule
\end{tabularx}
\caption{Frequency distributions of ratings regarding correctness of system answers for Exp.A and X.}
\label{tab:answer_correct_ExpAandB}
\end{table*}

\paragraph{Differences between Datasets and Models}

If we first look at Exp.A (Table~\ref{tab:ExpAX_results_grayscale_cond_allcriteria}), only minimal distinctions are evident between datasets or models, particularly concerning the models' ability to recognize colors, materials, and their overall competency.
While slight variations exist in the other evaluation criteria, none are notably remarkable. 
For instance, regarding their understanding of the general scene, the models exhibit slightly better performance with the CLEVR-X dataset.
In Exp.X (Table~\ref{tab:ExpAX_results_grayscale_cond_allcriteria}), on the other hand, the results exhibit some more variation between models and datasets. 
For example, only for the models' overall competency, do we find the same (median) value across models and datasets.  
Overall, it also appears that the items based on CLEVR-X data perform slightly better in Exp.X, specifically in terms of the models' ability to recognize shapes and materials, as well as their general scene understanding and overall competence.

Table~\ref{tab:color-term-in-question-distribution} shows the frequency of questions in the human evaluation study that contain the word “color[s]” or specific color terms like “red” or “blue” etc., categorized by dataset.
 It is evident that almost all questions in the CLEVR-X dataset contain color terms, with about half explicitly mentioning the word “color”. Conversely, in the VQA-X dataset, only three out of 64 questions include the word “color[s]”. Hence, the observed distinctions between the datasets may be attributed to this contrast.

\begin{table*}[!htbp]
\centering

\small
\begin{tabularx}{\linewidth}{lYYYY}
\toprule
        & \multicolumn{2}{c}{“Color[s]” in question} & \multicolumn{2}{c}{Color term in question} \\
        \cmidrule(lr){2-3} \cmidrule(lr){4-5}
Dataset & yes & no & yes & no \\
\midrule
    CLEVR-X & 34 & 30 & 59 &  5 \\
    VQA-X   &  3 & 61 &  3 & 61 \\
\bottomrule
\end{tabularx}

\caption{Occurrence of questions in the human evaluation study containing the word “color[s]” or specific color terms like “red” or “blue”, differentiated by dataset (color terms include any instance of “color”, a specific color term, or both).}
\label{tab:color-term-in-question-distribution}
\end{table*}

\begin{table*}[t]
\small
\centering
\begin{tabularx}{\textwidth}{lllYYYYYY}
\toprule
    &      &     & \multicolumn{2}{c}{Consist. of Expl. \& Answ.} & \multicolumn{2}{c}{Consist. of Expl. \& Img.} & \multicolumn{2}{c}{Fluency of Expl.} \\
    \cmidrule(lr){4-5}\cmidrule(lr){6-7}\cmidrule(lr){8-9}
    Condition & Dataset & Model &         median &  mean &        median &  mean &      median &  mean \\
\midrule
\textbf{grayscale} & CLEVR-X  & NLX-GPT &            4.0 & \textbf{ 3.26} &           1.0 &  1.53 &         4.0 &  3.27 \\
    &      & Uni-NLX &            4.0 &  3.17 &           1.0 &  1.74 &         4.0 &  \textbf{3.46} \\
    &  VQA-X & NLX-GPT &            2.0 &  2.67 &           1.0 &  1.85 &         4.0 &  3.42 \\
    &      & PJ-X &            1.0 &  2.20 &           1.0 &  \textbf{2.02} &         4.0 &  3.35 \\
\midrule
\textbf{color} & CLEVR-X  & NLX-GPT &            5.0 &  4.58 &           5.0 &  4.53 &         5.0 &  4.52 \\
    &      & Uni-NLX &            5.0 &  \textbf{4.61 }&           5.0 &  \textbf{4.59 }&         5.0 &  \textbf{4.54} \\
    &  VQA-X & NLX-GPT &            5.0 &  4.42 &           5.0 &  4.53 &         5.0 &  4.34 \\
    &      & PJ-X &            4.0 &  3.56 &           4.0 &  3.63 &         5.0 &  3.85 \\
\bottomrule
\end{tabularx}
\caption{Human ratings for the additional evaluation criteria of \textbf{Exp.X}. We asked the participants to rate the \emph{consistency of the explanation with the answer}, the \emph{consistency of the explanation with the image}, and the \emph{fluency of the explanation}. We report the median and mean scores across raters as the final scores, with bold values indicating
conditions with the best (mean) values for that evaluation criteria.}
\label{tab:ExpB_results_additionalcriteria}
\end{table*}

\begin{figure*}[h]
    \centering
    \begin{subfigure}{0.49\columnwidth}
        \centering
        \includegraphics[width=\linewidth]{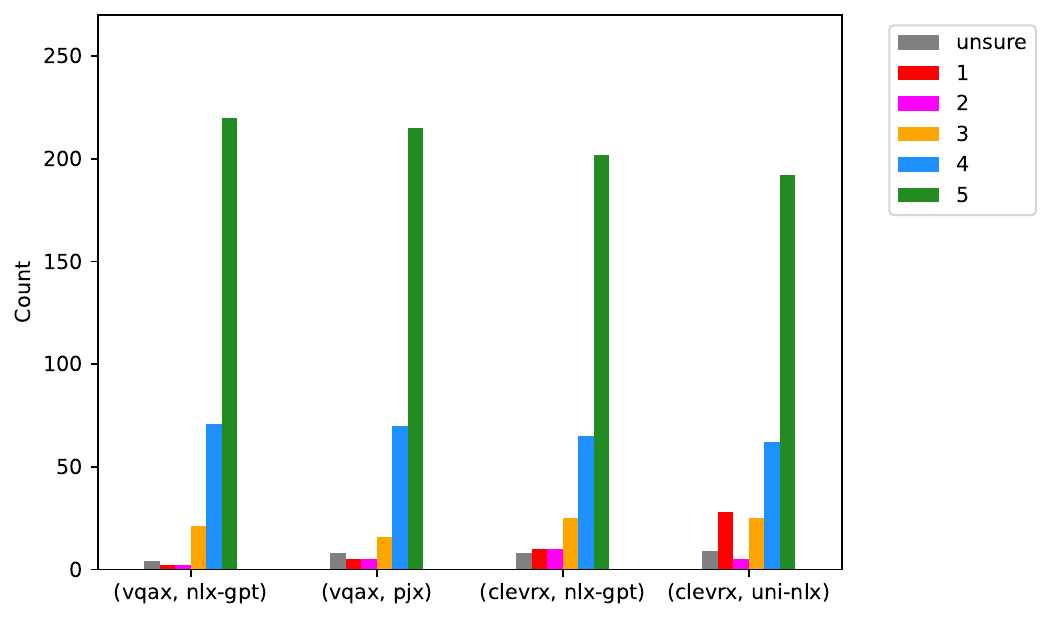}
        \caption{Exp.A -- colored images}
    \end{subfigure}
    \hfill
    \begin{subfigure}{0.49\columnwidth}
        \centering
        \includegraphics[width=\linewidth]{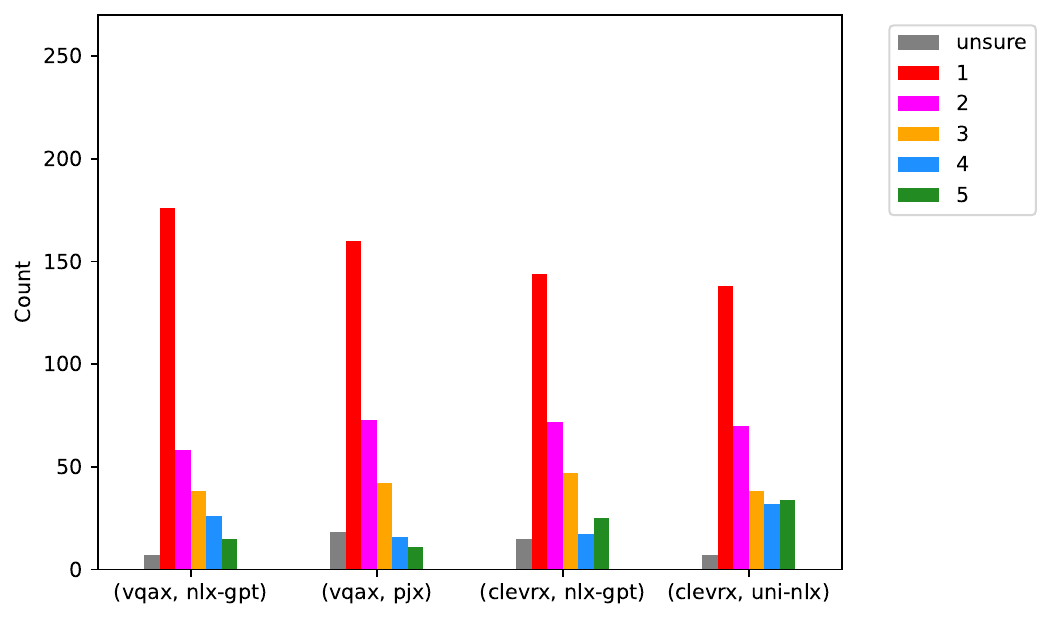}
        \caption{Exp.A -- grayscale images}
    \end{subfigure}
    \hfill
        \begin{subfigure}{0.49\columnwidth}
        \centering
        \includegraphics[width=\linewidth]{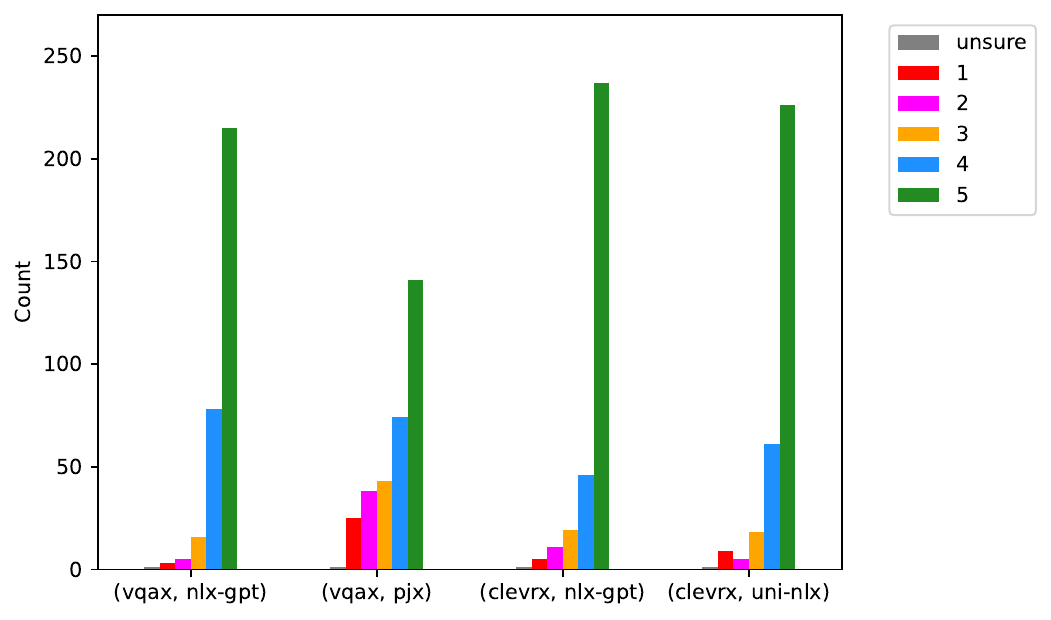}
        \caption{Exp.X -- colored images}
    \end{subfigure}
    \hfill
    \begin{subfigure}{0.49\columnwidth}
        \centering
        \includegraphics[width=\linewidth]{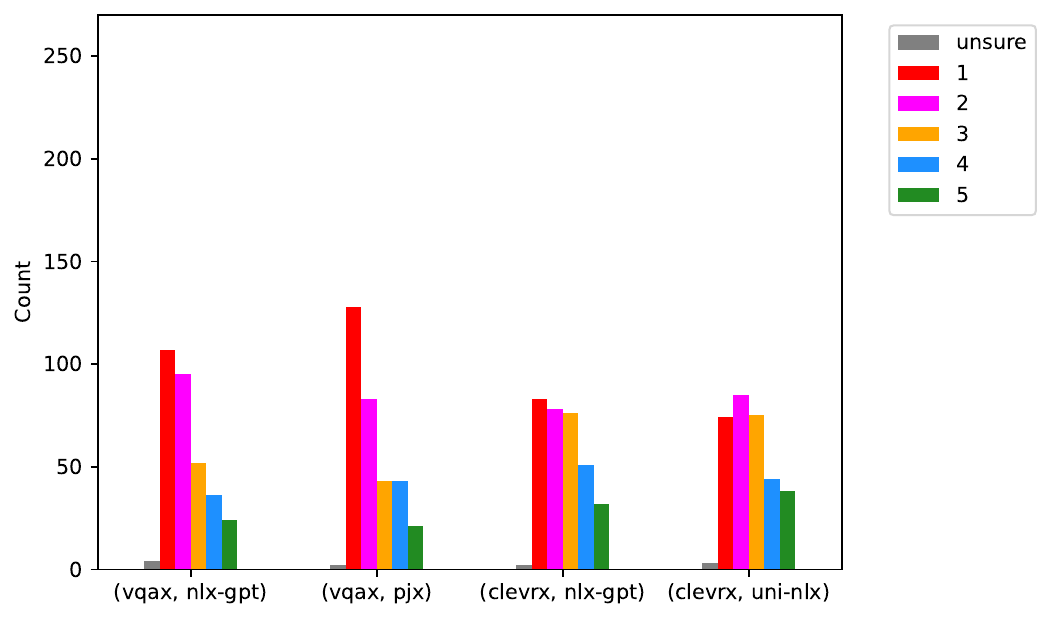}
        \caption{Exp.X -- grayscale images}
    \end{subfigure}
    \caption{Human ratings on the evaluation criterion “Ability of the AI system to \textbf{understand the general scene}”. Participants indicated their judgment on a scale from 1 (strongly disagree; here in red) to 5 (strongly agree; here in green).}  
    \label{fig:plots_genscene}
\end{figure*}

\begin{figure*}[h]
    \centering
    \begin{subfigure}{0.49\columnwidth}
        \centering
        \includegraphics[width=\linewidth]{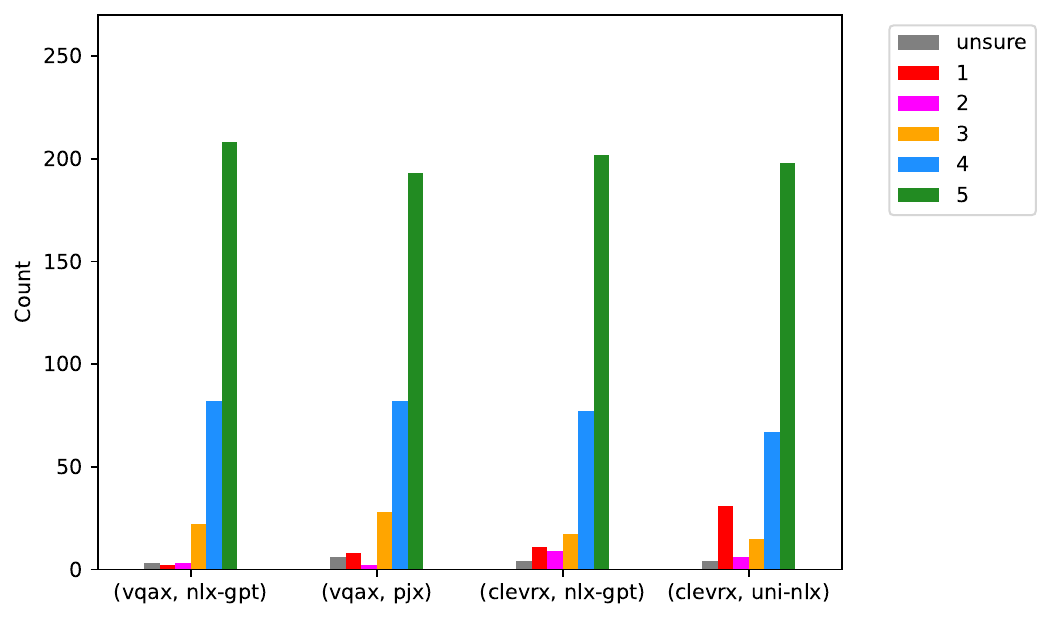}
        \caption{Exp.A -- colored images}
    \end{subfigure}
    \hfill
  \begin{subfigure}{0.49\columnwidth}
        \centering
        \includegraphics[width=\linewidth]{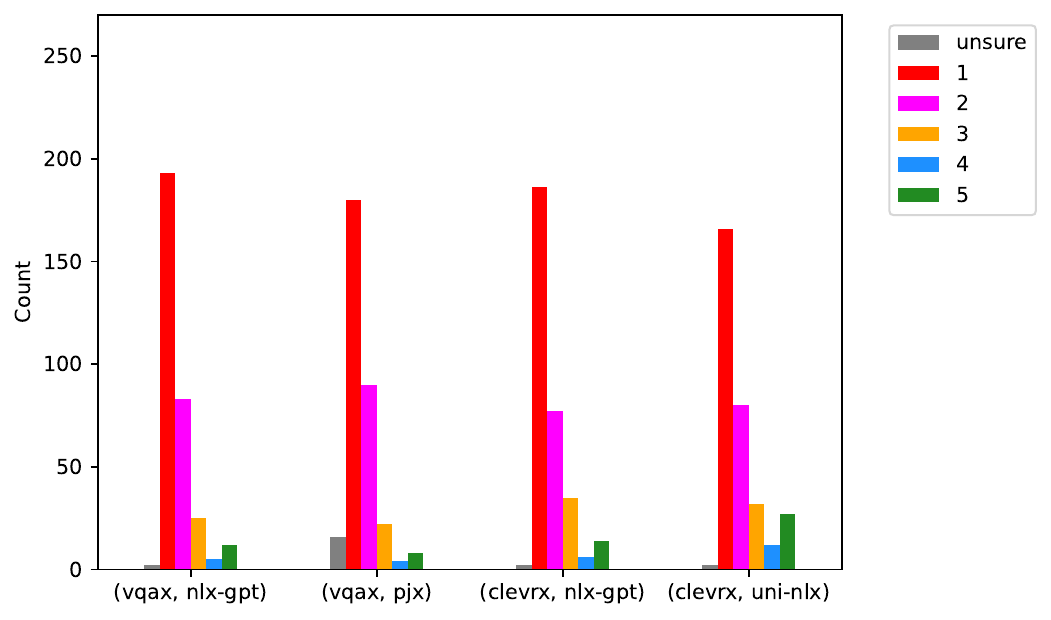}
        \caption{Exp.A -- grayscale images}
    \end{subfigure}
    \hfill
    \begin{subfigure}{0.49\columnwidth}
        \centering
        \includegraphics[width=\linewidth]{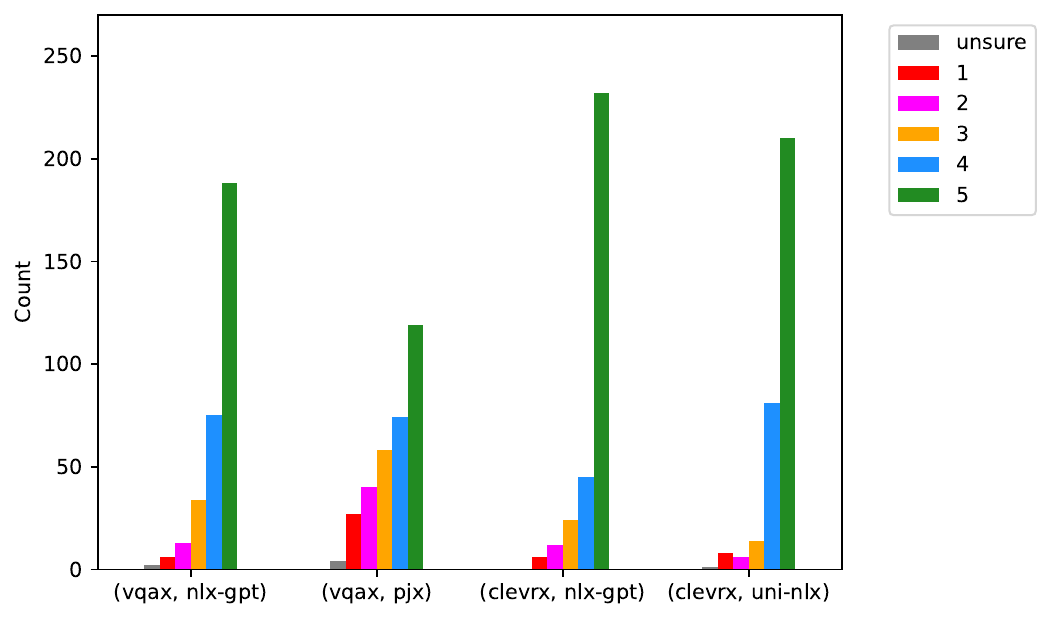}
        \caption{Exp.X -- colored images}
    \end{subfigure}
    \hfill
    \begin{subfigure}{0.49\columnwidth}
        \centering
        \includegraphics[width=\linewidth]{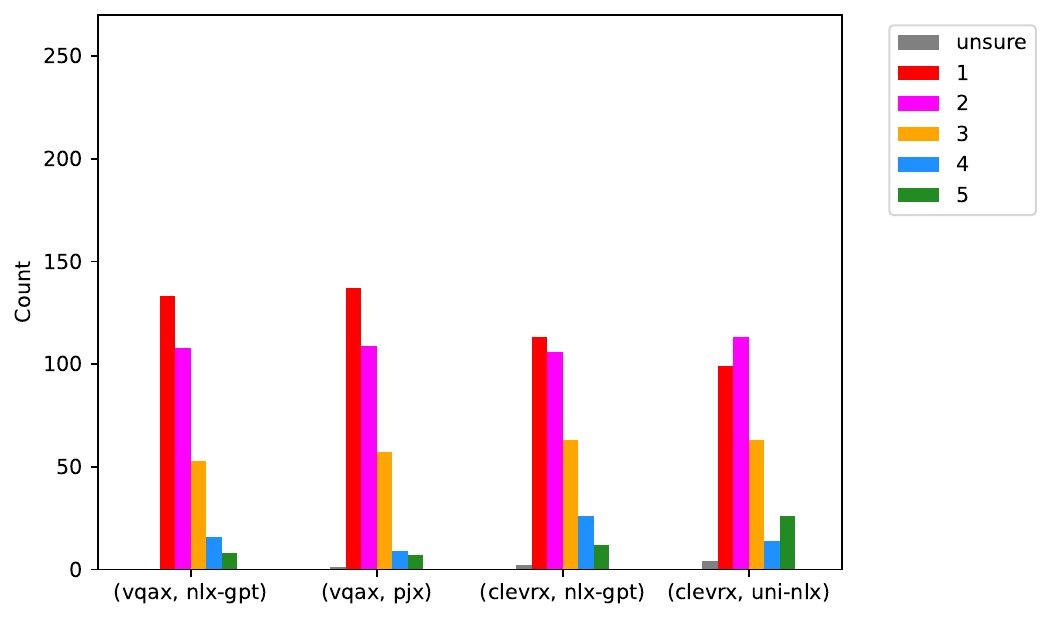}
        \caption{Exp.X -- grayscale images}
    \end{subfigure}
    
    \caption{Human ratings on the evaluation criterion “\textbf{Overall competency} of the AI system”. Participants indicated their judgment on a scale from 1 (strongly disagree; here in red) to 5 (strongly agree; here in green).}  
    \label{fig:plots_competency}
\end{figure*}

\begin{figure*}[h]
    \centering
    \begin{subfigure}{0.49\columnwidth}
        \centering
        \includegraphics[width=\linewidth]{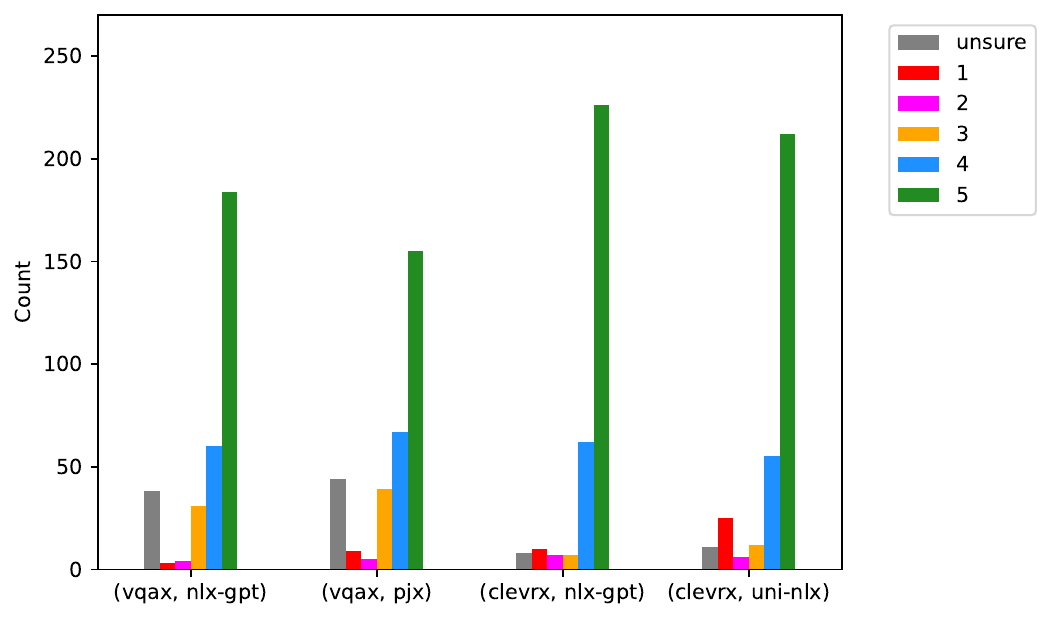}
        \caption{Exp.A -- colored images}
    \end{subfigure}
    \hfill    
  \begin{subfigure}{0.49\columnwidth}
        \centering
        \includegraphics[width=\linewidth]{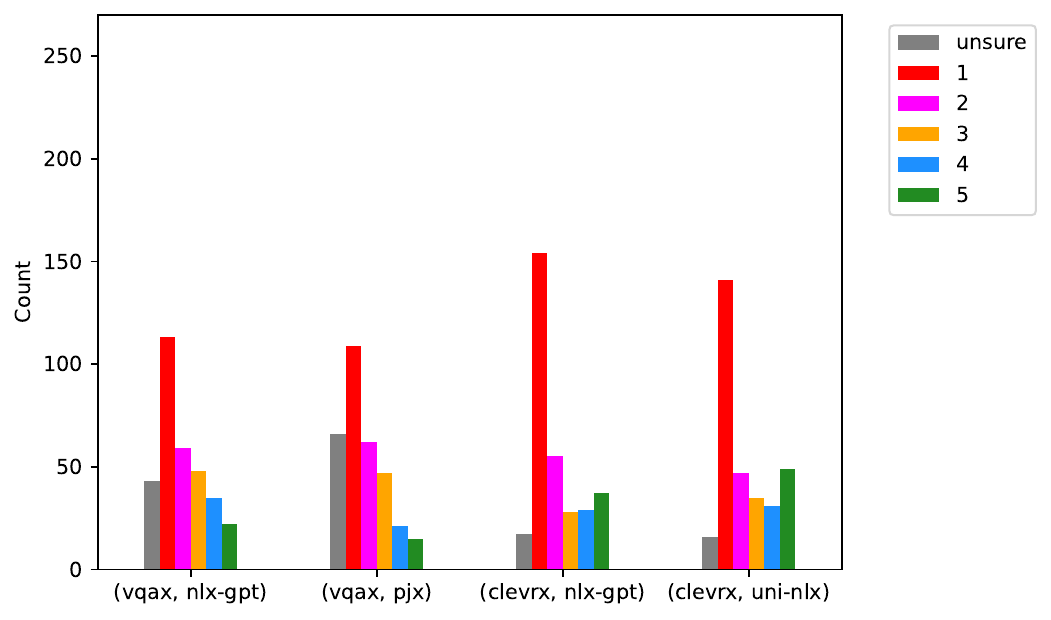}
        \caption{Exp.A -- grayscale images}
    \end{subfigure}
    \hfill
    \begin{subfigure}{0.49\columnwidth}
        \centering
        \includegraphics[width=\linewidth]{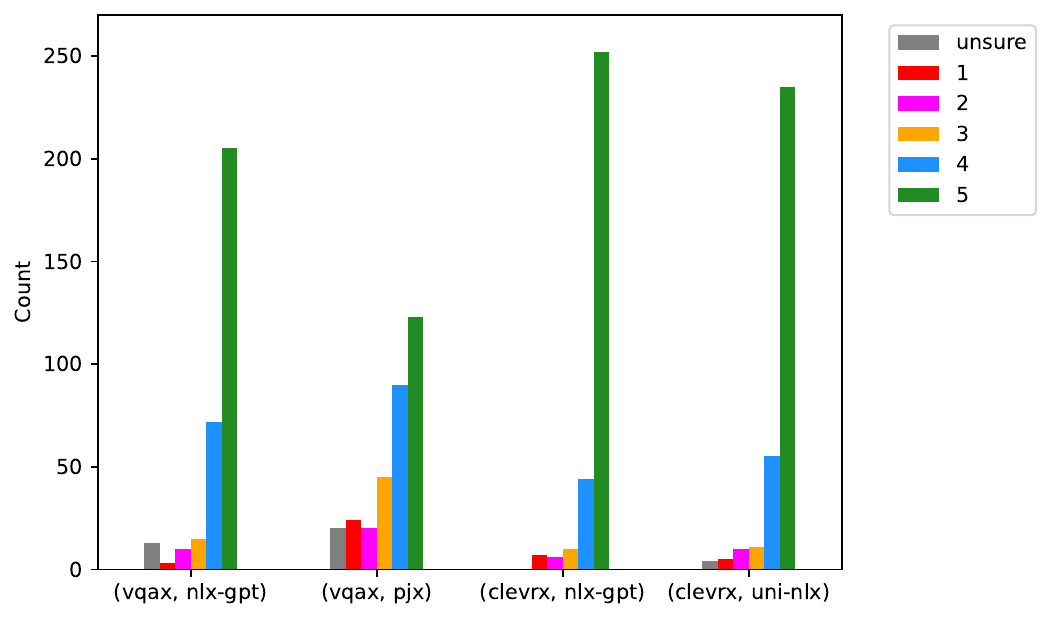}
        \caption{Exp.X -- colored images}
    \end{subfigure}
    \hfill
    \begin{subfigure}{0.49\columnwidth}
        \centering
        \includegraphics[width=\linewidth]{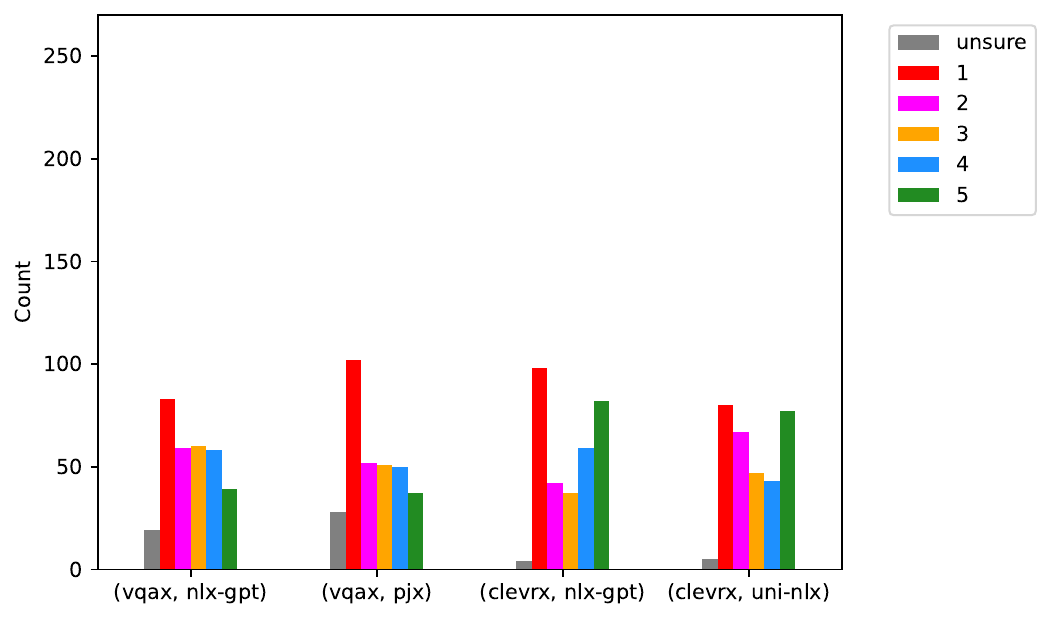}
        \caption{Exp.X -- grayscale images}
    \end{subfigure}
    \caption{Human ratings on the evaluation criterion “Ability of the  AI system to \textbf{recognize shapes}”. Participants indicated their judgment on a scale from 1 (strongly disagree; here in red) to 5 (strongly agree; here in green).}  
    \label{fig:plots_shapes}
\end{figure*}

\begin{figure*}[h]
    \centering
    \begin{subfigure}{0.49\columnwidth}
        \centering
        \includegraphics[width=\linewidth]{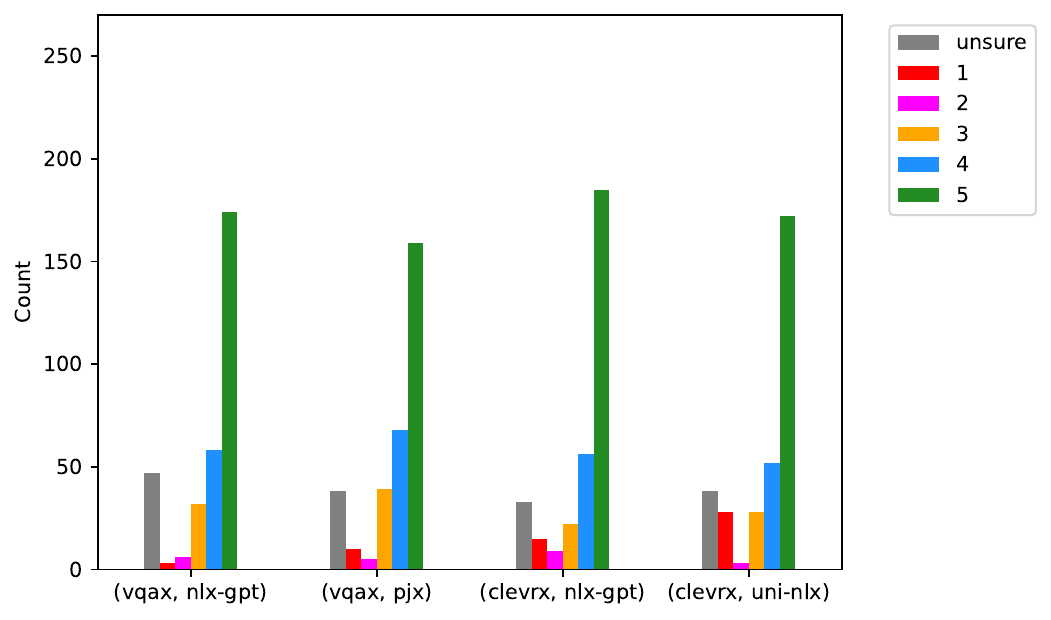}
        \caption{Exp.A -- colored images}
    \end{subfigure}
  \hfill    
  \begin{subfigure}{0.49\columnwidth}
        \centering
        \includegraphics[width=\linewidth]{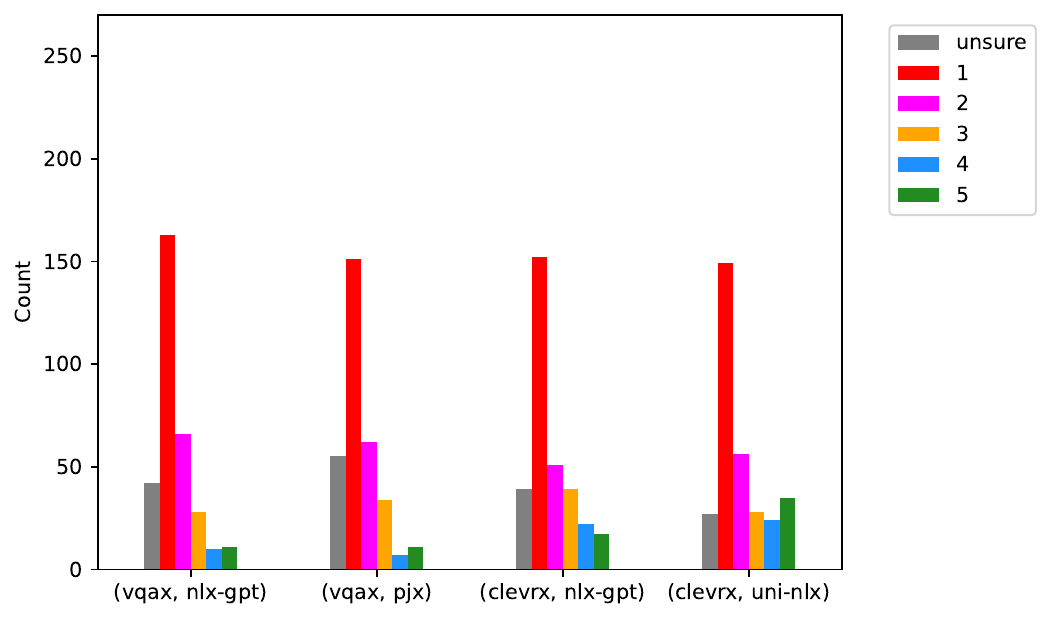}
        \caption{Exp.A -- grayscale images}
    \end{subfigure}
    \hfill
    \begin{subfigure}{0.49\columnwidth}
        \centering
        \includegraphics[width=\linewidth]{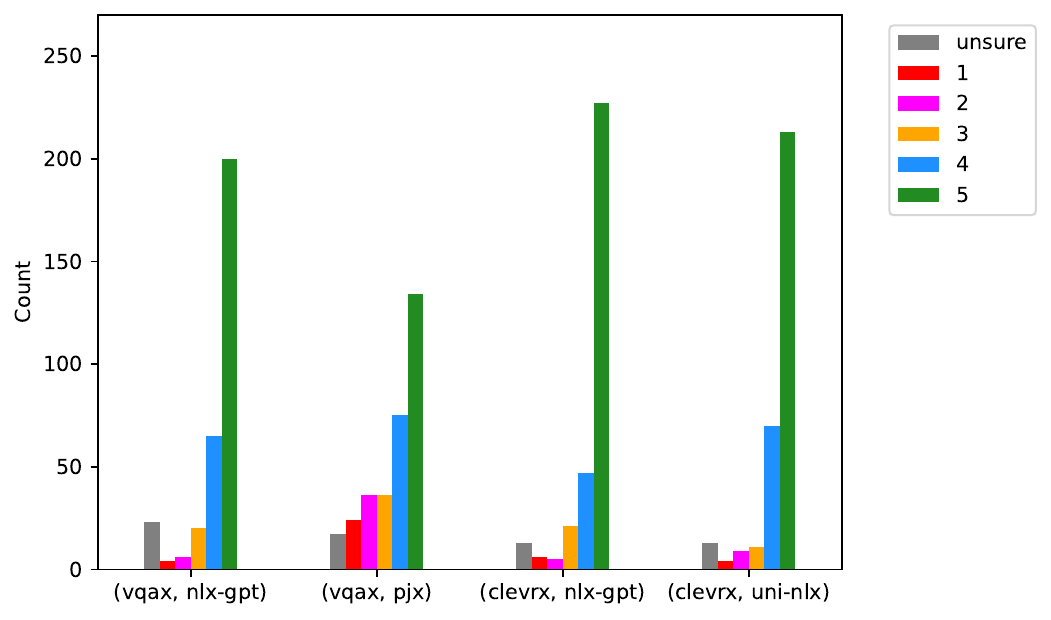}
        \caption{Exp.X -- colored images}
    \end{subfigure}
    \hfill
    \begin{subfigure}{0.49\columnwidth}
        \centering
        \includegraphics[width=\linewidth]{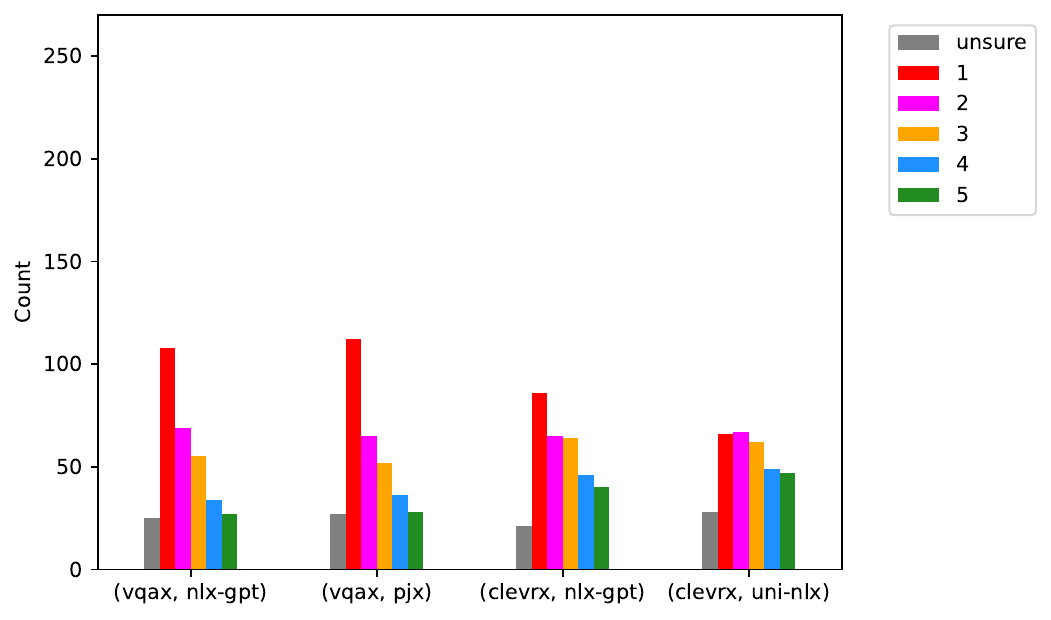}
        \caption{Exp.X -- grayscale images}
    \end{subfigure}
    
    \caption{Human ratings on the evaluation criterion “Ability of the  AI system to \textbf{recognize materials}”. Participants indicated their judgment on a scale from 1 (strongly disagree; here in red) to 5 (strongly agree; here in green).}  
    \label{fig:plots_materials}
\end{figure*}

\paragraph{Analysis of the Color Condition}

Table~\ref{tab:ExpAX_results_color_cond_allcriteria} shows the human evaluation results for the color condition in Exp. A and X.
In contrast to the results of the grayscale condition (Table~\ref{tab:ExpAX_results_grayscale_cond_allcriteria}), with respect to all the evaluation criteria, the evaluation for both Exp.A and Exp.X is very good. This corresponds to our expectation because  only items with correct model answers were included in the color condition.

Furthermore, we can see that in both Exp.A and Exp.X, there are no remarkable differences between the ability to recognize colors and the other tested abilities. This is also evident from the Mann-Whitney U Test results in Table~\ref{tab:color_mann-whitney-u-test}, especially when compared to the Mann-Whitney U results for the grayscale condition in Table~\ref{tab:grayscale_mann-whitney-u-test}.

However, it is notable that, with respect to all evaluation criteria, the PJ-X model receives lower ratings in Exp.X compared to Exp.A. In other words, including explanations in Exp.X results in a decline in performance for the PJ-X model.
For the other models, we do not observe this difference between the two Experiments; instead, their evaluation remains fairly consistent in the color condition across both experiments. 
Consequently, the explanations produced by the PJ-X model seem inferior to those of the other models. This discrepancy may be due to the unique architecture of the PJ-X model, which, unlike the other models, generates answers and explanations in two separate steps rather than one.

\begin{table*}[h]
\small
\centering
\begin{tabularx}{\textwidth}{lllYYYYYYYYYY}
\toprule
     & &     & \multicolumn{2}{c}{Colors} & \multicolumn{2}{c}{Shapes} & \multicolumn{2}{c}{Materials} & \multicolumn{2}{c}{General Scene} & \multicolumn{2}{c}{Competency} \\
     \cmidrule(rl){4-5} \cmidrule(rl){6-7} \cmidrule(rl){8-9} \cmidrule(rl){10-11} \cmidrule(rl){12-13}
Experiment & Dataset     & Model     & med &  mean & med &  mean &    med &  mean &    med &  mean &    med &  mean \\
\midrule
\textbf{Exp. A.} & CLEVR-X  & NLX-GPT &    5.0 &  \textbf{4.55 }&    5.0 &  \textbf{4.57} &       5.0 &  4.34 &       5.0 &  4.43 &       5.0 &  4.47 \\
     & & Uni-NLX &    5.0 &  4.33 &    5.0 &  4.38 &       5.0 &  4.20 &       5.0 &  4.23 &       5.0 &  4.28 \\
    & VQA-X & NLX-GPT &    5.0 &  \textbf{4.55 }&    5.0 &  4.50 &       5.0 &  \textbf{4.45 }&       5.0 &  \textbf{4.67} &       5.0 &  \textbf{4.66} \\
    & & PJ-X &    5.0 &  4.38 &    5.0 &  4.30 &       5.0 &  4.30 &       5.0 &  4.57 &       5.0 &  4.50 \\
\midrule
\textbf{Exp.X} & CLEVR-X  & NLX-GPT &    5.0 &  4.65 &    5.0 &  \textbf{4.66} &       5.0 &  \textbf{4.58} &       5.0 &  4.57 &       5.0 &  4.52 \\
     & & Uni-NLX &    5.0 &  \textbf{4.74} &    5.0 &  4.61 &       5.0 &  4.56 &       5.0 & \textbf{ 4.58 }&       5.0 &  \textbf{4.56} \\
     & VQA-X & NLX-GPT &    5.0 &  4.54 &    5.0 &  4.54 &       5.0 &  4.54 &       5.0 &  \textbf{4.58} &       5.0 &  4.38 \\
     & & PJ-X &    4.0 &  3.80 &    4.0 &  3.86 &       4.0 &  3.84 &       4.0 &  3.86 &       4.0 &  3.71 \\
\bottomrule
\end{tabularx}
\caption{Human ratings on the different evaluation criteria for the \textbf{color condition} of Exp.A (i.e., no model explanations were shown to the participants) and Exp.B (i.e., model explanations were shown to the participants). For \emph{Colors}, \emph{Shapes} and \emph{Materials}, we asked the participants to rate the AI system's ability to recognize the respective capability. Further, we asked the participants to rate the AI system's understanding of the \emph{General Scene} as well as it's overall \emph{Competency}. We report the median and mean scores across raters as the final scores. Bold values indicate conditions with the best (mean) values for that evaluation criteria.}
\label{tab:ExpAX_results_color_cond_allcriteria}
\end{table*}

\paragraph{Correlations between BERTscore and human judgments}

Table~\ref{tab:BERTScore-correlation-human-and-automatic-metrics} shows Pearson’s correlation coefficients ($\rho$) between the automatic and human evaluation metrics for the CLEVR-X and VQA-X datasets. Interestingly, we find large differences between the datasets. While all human metrics show statistically significant correlations with BERTScore for the VQA-X dataset, we find no statistically significant correlations for the CLEVR-X dataset.
However, one commonality between the two datasets is the lack of differentiation between various criteria. The fact that all skills either correlate or show no correlation suggests that the automatic BERTScore metric is not able to capture the nuanced distinctions that human evaluation can discern.

\begin{table*}[!htbp]
\centering
\small
\begin{tabularx}{\linewidth}{llYYYY}
\toprule
                 &              & \multicolumn{2}{c}{CLEVR-X} & \multicolumn{2}{c}{VQA-X} \\
                 \cmidrule(rl){3-4} \cmidrule(rl){5-6}
Automatic metric & Human metric & $\rho$ & $p$-value & $\rho$ & $p$-value \\
\midrule
    \multirow{8}{*}{BERTScore}
    & Consist. of Expl. \& Answ. & -0.090 & 0.31 & 0.251 & \textbf{0.008} \\
    & Consist. of Expl. \& Img.  & -0.020 & 0.82 & 0.278 & \textbf{0.003} \\
    & Fluency of Expl.           & -0.033 & 0.71 & 0.304 & \textbf{0.001} \\
    & Shapes                     & -0.068 & 0.44 & 0.231 & \textbf{0.02}  \\
    & Colors                     & -0.023 & 0.80 & 0.201 & \textbf{0.04}  \\
    & Materials                  & -0.056 & 0.53 & 0.248 & \textbf{0.009} \\
    & General Scene              & -0.051 & 0.57 & 0.251 & \textbf{0.008} \\
    & Competency                 & -0.051 & 0.57 & 0.252 & \textbf{0.008} \\
\bottomrule
\end{tabularx}
\caption{Pearson’s correlation coefficient ($\rho$) between BERTScore results and human evaluation metrics for CLEVR-X and VQA-X data. $p$-values in bold indicate statistical significance ($p < 0.05$).}
\label{tab:BERTScore-correlation-human-and-automatic-metrics}
\end{table*}

\subsection{Online Experiment}
\label{sec:appendix_online_experiment}

\begin{figure*}[!ht]
    \centering
    \includegraphics[width=0.8\linewidth]{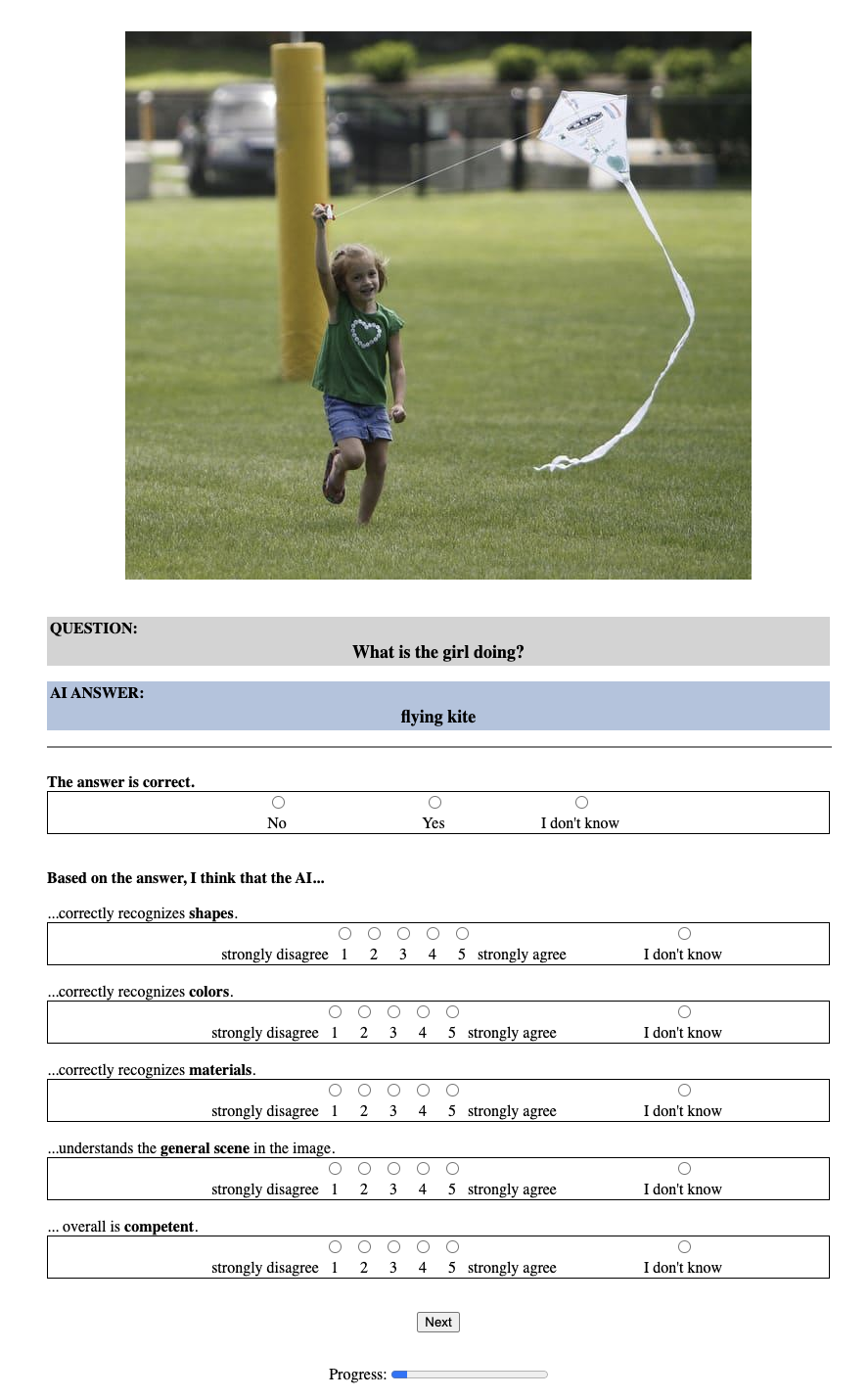}
    \caption{A training item used in the online experiment to familiarize participants with the task and rating scales. This item comes from the \textbf{VQA-X} dataset and from \textbf{Exp.A}, i.e., the study without explanations.}
    \label{fig:screenshot_expA_vqax_trialrun}
\end{figure*}

\begin{figure*}[!ht]
    \centering
    \includegraphics[width=0.9\linewidth]{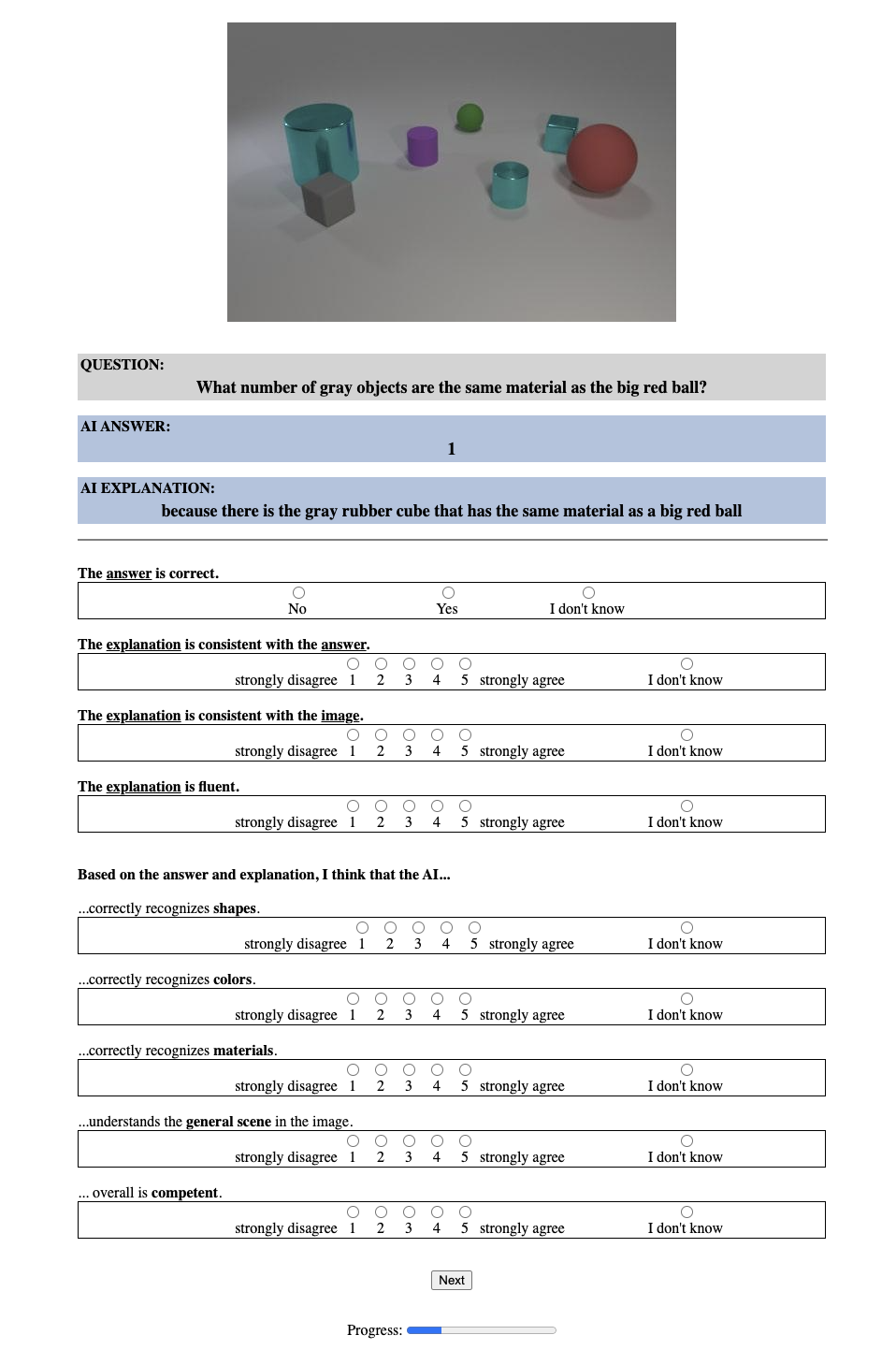}
    \caption{An experimental item used in the online experiment. This item comes from the \textbf{CLEVR-X} dataset and from \textbf{Exp.X}, i.e., the experiment with explanations.}
    \label{fig:screenshot_expX_cleverx_trialrun}
\end{figure*}

Figures~\ref{fig:screenshot_expA_vqax_trialrun} and \ref{fig:screenshot_expX_cleverx_trialrun} show screenshots of the study, example items and evaluation criteria.

\end{document}